%% file: main.tex
\documentclass[letterpaper]{article} 
\usepackage{aaai23}  
\usepackage{times}  
\usepackage{helvet}  
\usepackage{courier}  
\usepackage[hyphens]{url}  
\usepackage{graphicx} 
\usepackage{natbib}  
\usepackage{caption} 
\input{command/mycommand}
\input{command/math_command}

\newcommand{\alg}{\code{DENEB}}

\usepackage[utf8]{inputenc} 
\usepackage{url}            
\usepackage{booktabs}       
\usepackage{amsfonts}       
\usepackage{nicefrac}       
\usepackage{microtype}      
\usepackage{xcolor}         

\usepackage[linesnumbered,ruled,vlined]{algorithm2e}
\SetKwInput{KwInput}{Input}                
\SetKwInput{KwOutput}{Output}              

\SetCommentSty{mycommfont}

\title{Denoising after Entropy-based Debiasing \\ A Robust Training Method for Dataset Bias with Noisy Labels}
\author{
    Sumyeong Ahn,
    Se-Young Yun
}
\affiliations{
    Kim Jaechul Graduate School of AI, KAIST, Seoul, Republic of Korea

    \{sumyeongahn, yunseyoung\}@kaist.ac.kr
}

\begin{document}

\maketitle

\begin{abstract}
Improperly constructed datasets can result in inaccurate inferences. For instance, models trained on biased datasets perform poorly in terms of generalization (\ie \emph{dataset bias}). Recent debiasing techniques have successfully achieved generalization performance by underestimating easy-to-learn samples (\ie \emph{bias-aligned samples}) and highlighting difficult-to-learn samples (\ie \emph{bias-conflicting samples}). However, these techniques may fail owing to noisy labels, because the trained model recognizes noisy labels as difficult-to-learn and thus highlights them. In this study, we find that earlier approaches that used the provided labels to quantify difficulty could be affected by the small proportion of noisy labels. Furthermore, we find that running denoising algorithms before debiasing is ineffective because denoising algorithms reduce the impact of difficult-to-learn samples, including valuable bias-conflicting samples. Therefore, we propose an approach called denoising after entropy-based debiasing, \ie \alg, which has three main stages. (1) The prejudice model is trained by emphasizing (bias-aligned, clean) samples, which are selected using a Gaussian Mixture Model. (2) Using the per-sample entropy from the output of the prejudice model, the sampling probability of each sample that is proportional to the entropy is computed. (3) The final model is trained using existing denoising algorithms with the mini-batches constructed by following the computed sampling probability. Compared to existing debiasing and denoising algorithms, our method achieves better debiasing performance on multiple benchmarks.
\end{abstract}

\input{main/intro}

\input{main/problem}
\input{main/debias}
\input{main/denoise}
\input{main/method}
\input{main/exp}
\input{main/related}
\input{main/conclusion}

\section*{Acknowledgements}
This work was supported by Institute of Information \& communications Technology Planning \& Evaluation (IITP) grant funded by the Korea government(MSIT) (No.2019-0-00075, Artificial Intelligence Graduate School Program(KAIST), 10\%) and the Institute of Information \& communications Technology Planning \& Evaluation(IITP) grant funded by the Korea government(MSIT) (No. 2022-0-00871, Development of AI Autonomy and Knowledge Enhancement for AI Agent Collaboration, 90\%)
\bibliography{ref}

\input{appendix/appendix}

\end{document}

%% file: command/mycommand.tex
\usepackage{microtype}
\usepackage{subfig}
\usepackage{booktabs} 
\usepackage{bbm}
\usepackage{mathtools}
\usepackage{nccmath}
\usepackage{bm}

\usepackage{soul}
\usepackage{dsfont}
\usepackage{enumerate}
\usepackage{enumitem}

\usepackage{amsmath}
\usepackage{amsfonts}
\usepackage{bbm}
\usepackage{dsfont}
\usepackage[Symbol]{upgreek}
\usepackage{lscape}
\usepackage{caption}

\usepackage{wasysym}

\usepackage{multirow}
\usepackage{array, boldline, rotating}

\usepackage{kotex}

\usepackage{amssymb}
\usepackage{pifont}
\newcommand{\xmark}{\ding{55}}%

\newcommand{\mc}[1]{\mathcal{#1}}

\renewcommand*\eqref[1]{(\ref{#1})}

\newcommand{\eg}{\emph{e.g.,~}}
\newcommand{\ie}{\emph{i.e.,~}}

\newcommand{\myparagraph}[1]{\noindent\textbf{#1}~}

\def\code#1{\texttt{#1}}

\newcommand{\thickhline}{\hlineB{4}}
\newcommand{\bfcode}[1]{\code{\textbf{#1}}}

%% file: command/math_command.tex
\usepackage{amsmath,amsfonts,bm}

\def\eqref#1{equation~\ref{#1}}

\def\1{\bm{1}}

\DeclareMathAlphabet{\mathsfit}{\encodingdefault}{\sfdefault}{m}{sl}
\SetMathAlphabet{\mathsfit}{bold}{\encodingdefault}{\sfdefault}{bx}{n}

\DeclareMathOperator*{\argmax}{arg\,max}

%% file: main/intro.tex
\section{Introduction}
\label{sec:intro}

\begin{figure}
        \centering
        \includegraphics[width=1.0\columnwidth]{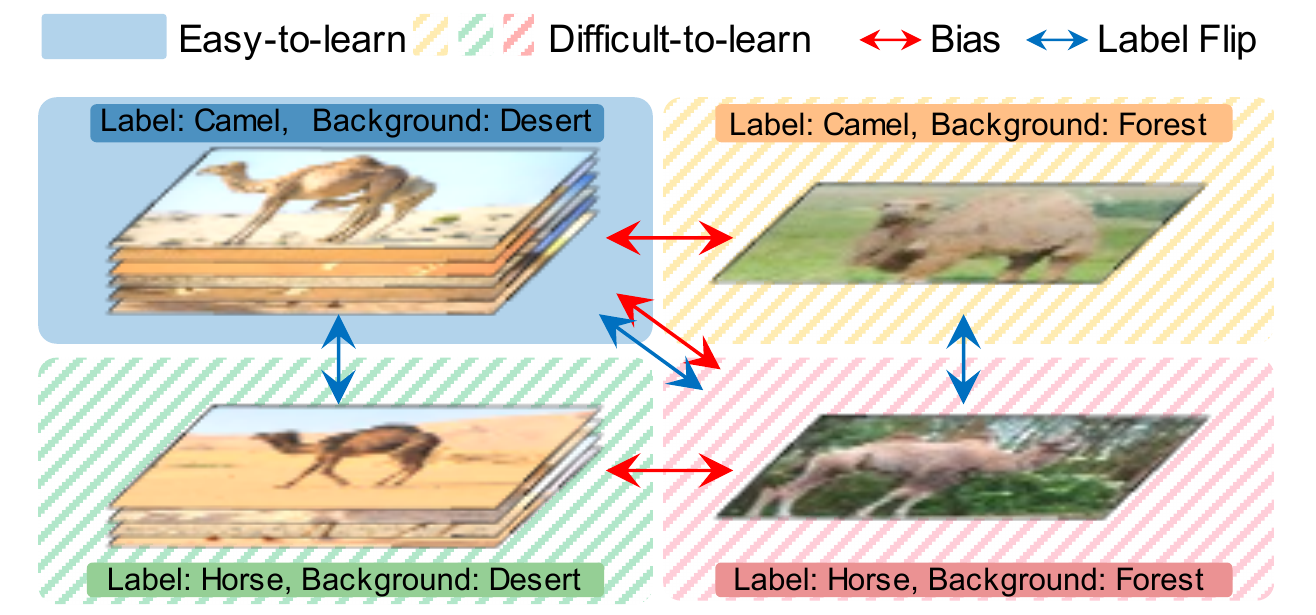}
        
        \caption{Examples of biased dataset with noisy labels. (1) \textcolor{blue}{Blue}: bias-aligned, clean label samples. (2) \textcolor{orange}{Orange}: bias-conflicting, clean labels samples. (3) \textcolor{green}{Green}: bias-aligned, noisy labels. (4) \textcolor{red}{Red}: bias-conflicting, noisy labels. The dashed background represents difficult-to-learn samples. Therefore, except for \textcolor{blue}{(bias-aligned, clean)} case, other cases are difficult-to-learn. To mitigate dataset bias with noisy labels, training directions for each type differ. For example, \textcolor{green}{(bias-aligned, noisy)} case must be discarded or cleansed, while \textcolor{orange}{(bias-conflicting, clean)} cases have to be emphasized.} 
        \label{fig:intro_ex}
\end{figure}

Deep neural networks (DNNs) have achieved human-like performance in various tasks, such as image classification~\cite{he2016deep}, image generation~\cite{goodfellow2014generative}, and object detection~\cite{he2017mask}, but require well-organized training datasets for success. For example, the trained model might have prejudices when its training dataset is biased. In real life, we often encounter \emph{dataset bias problems}~\cite{bahng2020learning, nam2020learning, lee2021learning, kim2021biaswap}. For example, as shown in Figure~\ref{fig:intro_ex}, the dataset for classifying camels in images could be very biased, as most camel images are captured against a desert background (\ie bias-aligned); only a few images are captured against other backgrounds, such as forests (\ie bias-conflicting). This unintended bias causes the trained model to infer erroneously based on shortcuts (\ie background). To mitigate such dataset bias, previous researches have used the fact that bias-conflicting samples are more difficult-to-learn than bias-aligned samples~\cite{nam2020learning}. Various approaches have been proposed, such as adjusting the loss function~\cite{bahng2020learning, nam2020learning, creager2021environment}, feature disentanglement~\cite{lee2021learning}, creating mixed-attribute samples~\cite{kim2021biaswap}, or reconstructing balanced datasets~\cite{liu2021just, ahn2021mitigating} to emphasize difficult-to-learn samples.

In addition to dataset bias, noisy labels are caused by many reasons~\cite{yu2018learning, nicholson2016label} and are known to degrade training mechanisms. For example, in Figure~\ref{fig:intro_ex}, the \emph{camel} in the bottom row can be labeled \emph{horse} by human error. Numerous studies have focused on alleviating the impact of noisy labels or directly correcting them. Some~\cite{bahri2020deep, zhang2020distilling, veit2017learning, ren2018learning, hendrycks2018using} deal with the problem of noisy labels by assuming clean data to set training guidelines (\ie purifying a given corrupted dataset by using a model trained on a small clean dataset). Recently, various methods have been proposed to relax this strong clean subset assumption by taking advantage of the characteristic that noisy labels are more difficult-to-learn than clean samples. For example, such difficult-to-learn samples are guided by regularizers~\cite{liu2020early, cao2019learning, cao2020heteroskedastic}, giving lower weights~\cite{wang2019symmetric, zhang2018generalized}, cleansing out~\cite{mirzasoleiman2020coresets, wu2020topological, pleiss2020identifying, han2018co, yu2019does}, or utilizing a semi-supervised learning algorithm by considering them as unlabeled samples~\cite{kim2021fine, li2019dividemix}.

Although dataset bias and noisy labels can occur simultaneously and independently, few studies~\cite{creager2021environment}\footnote{EIIL aimed to study dataset bias problem without human supervision. They partially analyzed the impact of noisy labels in their synthetic benchmark only.} have addressed both problems at once. This is because the fundamental solutions of each problem are exact opposites. Difficult-to-learn samples have to be \emph{emphasized} to mitigate dataset bias~\cite{nam2020learning, lee2021learning}, while their influence should be \emph{reduced} for denoising~\cite{han2018co, zhang2018generalized}. Dataset bias and noisy labels can occur concurrently in the real-world. Therefore, both problems must be handled together.

\begin{figure}
        \centering
        \includegraphics[width=1.0\columnwidth]{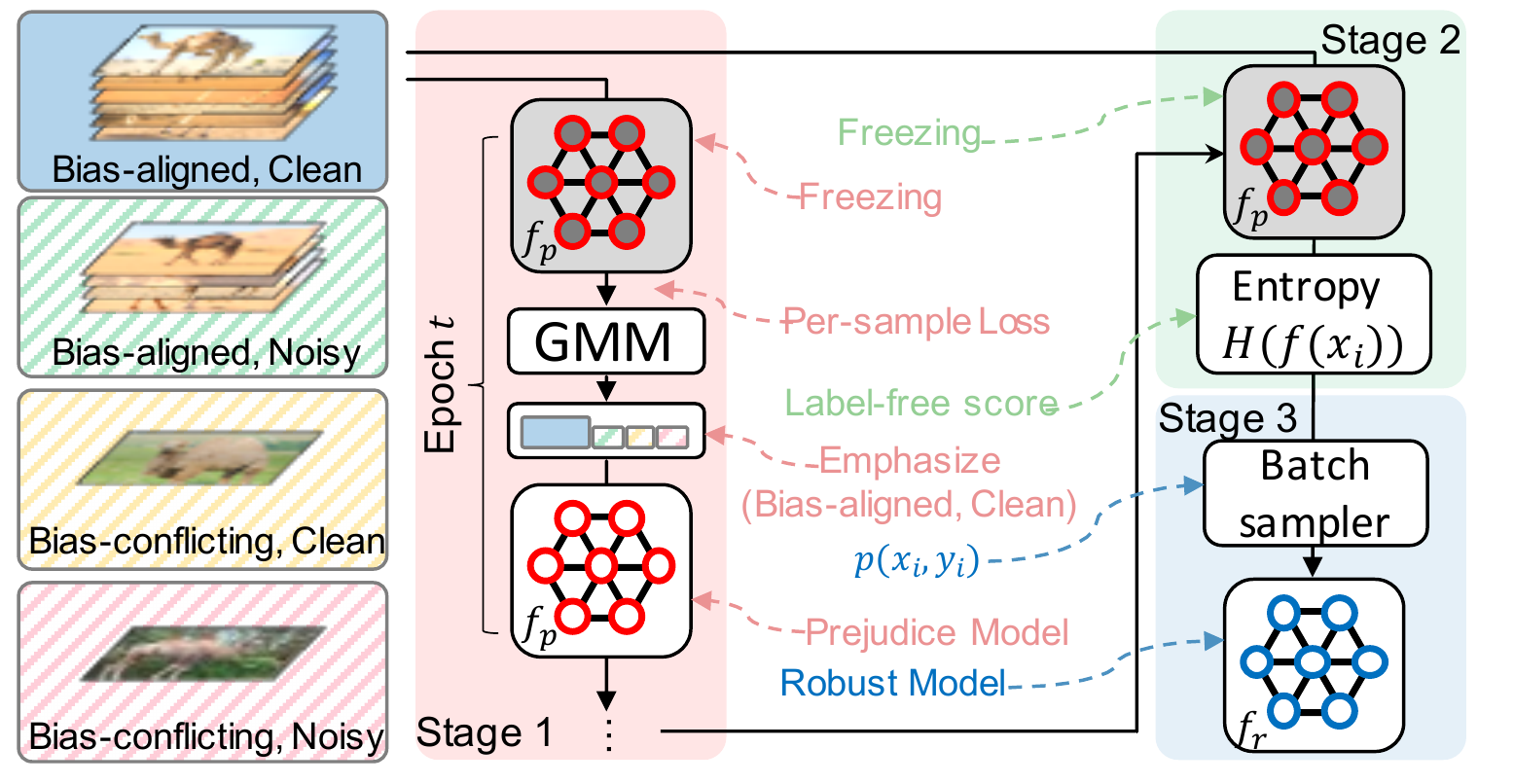}
        \caption{Overview of~\alg. It is composed of three steps. (1) Train by emphasizing \textcolor{blue}{(bias-aligned, clean)} samples. (2) Compute label-free score, \ie entropy. (3) Train the final robust model based on the batch sampler. }
        \label{fig:overview}
\end{figure}

\myparagraph{Contribution.} We present a training method that addresses \emph{dataset bias with noisy labels (DBwNL)}. We first empirically analyze why existing debiasing and denoising algorithms fail to achieve their respective objectives. In this regard, we discover two facts. (1) Existing debiasing methods using \emph{given labels} suffer from problems when the training dataset contains noisy labels because they determine the degree of being emphasized based on a given label (e.g., per-sample loss). 
(2) Denoising methods eliminate valuable bias-conflicting samples (\ie samples that should be emphasized for debiasing). This is because their denoising mechanism  cleans or discards difficult-to-learn samples without considering whether a sample is bias-conflicting or noisy label.


Based on these findings, we propose an algorithm, coined as \alg, which \textbf{\underline{den}}oising after \textbf{\underline{e}}ntropy-based de\textbf{\underline{b}}iasing (see Figure~\ref{fig:overview}). The proposed method consists of three stages. The first stage  trains a prejudice model biased toward \textcolor{blue}{(bias-aligned, clean)} data.
To this aim, \alg~ uses the Gaussian Mixture Model (GMM) based on per-sample losses to split \textcolor{blue}{(bias-aligned, clean)} and the others at the beginning of each epoch. In the next stage, \alg~ measures the entropy for each sample from the prejudice model. 
From the per-sample entropy, \alg~ calculates the sampling probability of each sample in proportion to the entropy. The key intuition of the sampling probability is that  \textcolor{green}{(bias-aligned, noisy)} samples are predicted to have low entropy by the prejudice model because it mainly learns \textcolor{blue}{(bias-aligned, clean)} samples, but the excluded bias-conflicting samples will have a large entropy prediction. Note that the sampling probability is obtained without using the given labels as they might be corrupted. Finally, \alg~ trains the ultimate robust model on the sampled mini-batches, where the sampling probabilities are obtained during Step 2.


We demonstrate the efficacy of \alg~on a variety of biased datasets, including Colored MNIST~\cite{nam2020learning, bahng2020learning, kim2021biaswap, lee2021learning}, Corrupted CIFAR-10~\cite{nam2020learning, lee2021learning}, Biased Action Recognition (BAR)~\cite{nam2020learning, kim2021biaswap}, and Biased Flickr-Faces-HQ~\cite{lee2021learning, kim2021biaswap}, with symmetric noisy labels. Compared to the existing debiasing, denoising, and naive combination of both algorithms, the proposed method achieves a successful debiasing performance for all benchmarks. For example, \alg~ improves the unbiased test accuracy from $39.24\%$ to $91.81\%$ on a colored MNIST dataset with $1\%$ bias scenario with $10\%$ noisy labels and BAR from $54.37\%$ to $62.30\%$ compared to the vanilla model.

%% file: main/problem.tex
\section{Dataset Bias with Noisy Labels (DBwNL)}
\label{sec:prob}

In this section, we define the dataset bias problem and the noisy label separately. Subsequently, we describe \emph{dataset bias with noisy labels}, which is when dataset bias and noisy label problems occur in conjunction.

\myparagraph{Dataset Bias.} Consider a dataset $\mc{D}=\{(x_i,y_i)\}_{i=1}^N$ in which each input is $x_i$ and its corresponding truth label $y_i=\{1,...,C\}$. Each sample can be described by a set of attributes. For example, the images in Figure~\ref{fig:intro_ex} can have \code{background}, \code{object}, and so on. For convenience of explanation, we look at the top of Figure~\ref{fig:intro_ex} (\textcolor{blue}{Blue} and \textcolor{orange}{Orange} cases),  without the noisy label case. The objective, \ie the \emph{target attribute}, is to classify the ``camel.'' 
When most of the samples have attributes that are strongly correlated with the target, we call the phenomenon \emph{dataset bias} and these attributes \emph{bias attributes}. In Figure~\ref{fig:intro_ex}, the bias-attribute ``desert background,'' and the target attribute ``camel'' are highly correlated, \ie almost ``camel'' images are captured against ``desert background.'' We call samples whose bias attribute is highly correlated (weakly correlated) with the target attribute \emph{bias-aligned} (\emph{bias-conflicting}) samples. This dataset bias problem is quite harmful when the bias attribute is easier to learn than the target attributes, because the model loses the motivation to learn the target attribute given its sufficiently low loss. We focus on the case where the bias attribute is easier to learn than the target. For convenience, we denote the set of bias-conflicting and bias-aligned samples by $\mc{D}_{c}$ and $\mc{D}_{a}$ as they are clearly distinct, but both sets need not be strictly separable. Note that the portion of bias-conflicting sample is called the bias conflict ratio $\alpha$, which is defined as: 
\begin{equation*}
    \alpha = \frac{|\mc{D}_c|}{|\mc{D}|}.
\end{equation*}

\myparagraph{Noisy Labels.} Collected labels may be corrupted. If a person labels the image $x$, the provided label can be corrupted \ie $y_\text{given} \neq y$, even though the true label is $y$. We call samples whose labels are $y_\text{given} = y$ and $y_\text{give} \neq y$ \emph{clean label} and \emph{noisy label}, respectively. As shown in Figure~\ref{fig:intro_ex}, the lower row represents noisy label cases. For example, although the images in Figure~\ref{fig:intro_ex} of the bottom boxes are ``camel'', they are labeled as ``horse''. We denote the portion of the samples whose labels are flipped as the noise ratio $\eta$. For convenience, we focus on the cases where the label corruption occurs symmetrically.
\begin{equation*}
    y_{\text{given}} = 
    \begin{cases}
      \tilde{y}\sim\text{Uniform}(C). & \text{with probability } \eta\\
      y & \text{with probability } 1-\eta
    \end{cases},
\end{equation*}

\myparagraph{DBwNL.}
DBwNL cases occur sequentially, gathering images $x$ and labeling $y$. As mentioned above, the biased dataset has a small portion of bias-conflicting samples, \ie $\alpha$ is small. Therefore, most of the samples in the given training dataset are bias-aligned. Training a robust model on the DBwNL dataset emphasizes bias-conflicting samples while discarding or reducing the impact of the noisy labels.

%% file: main/debias.tex
\section{Failure to debias on a DBwNL dataset}
\label{sec:debiasing}

In this section, we briefly summarize existing debiasing methods and demonstrate that they are vulnerable to noisy labels. 

\begin{figure}[t!]
    \centering
        \centering
        \subfloat[\code{LfF}~\cite{nam2020learning}]{\label{fig:lff}\includegraphics[width=0.49\linewidth]{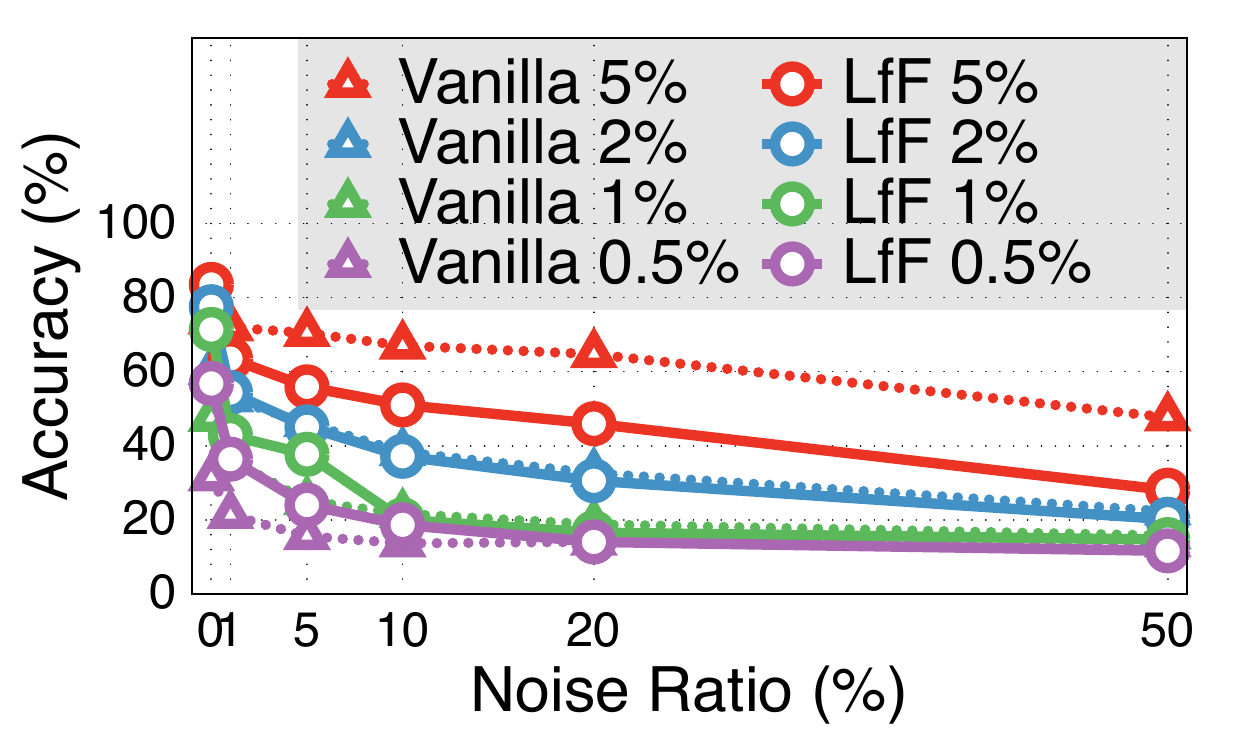}}
        \subfloat[\code{JTT}~\cite{liu2021just}]{\label{fig:jtt}\includegraphics[width=0.49\linewidth]{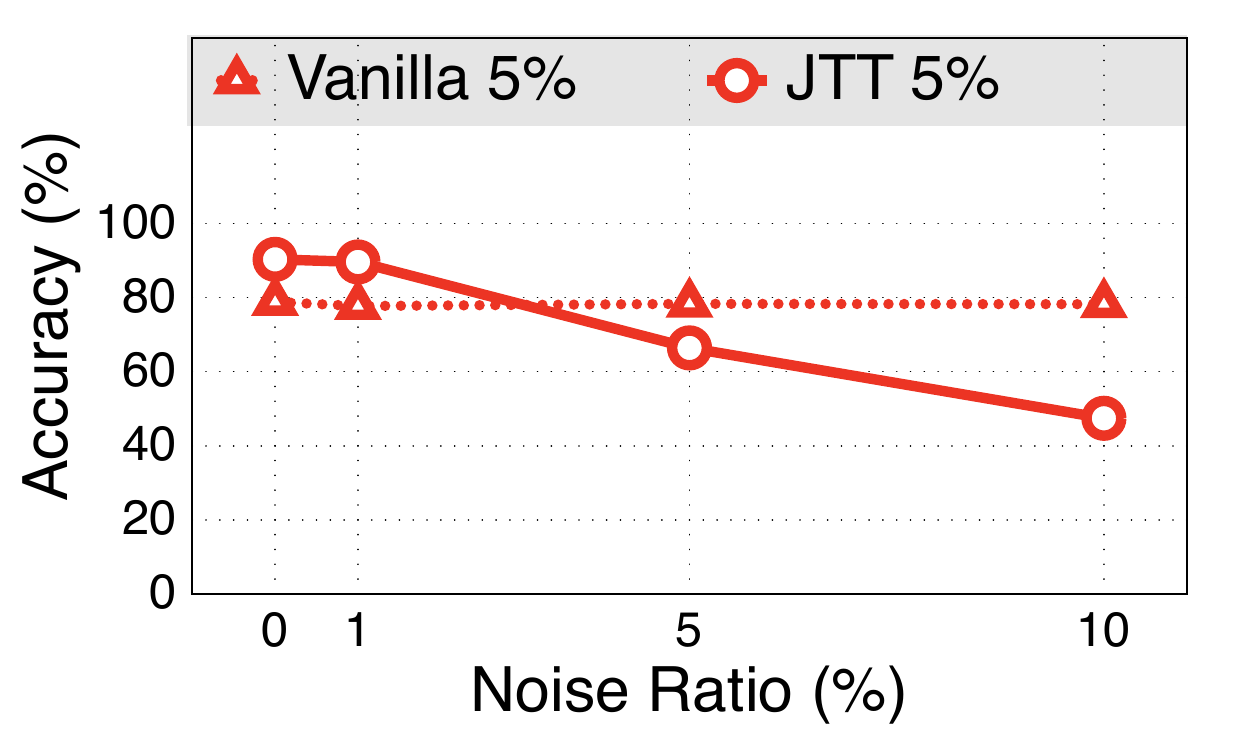}}\\
        \subfloat[\code{Disen}~\cite{lee2021learning}]{\label{fig:disen}\includegraphics[width=0.49\linewidth]{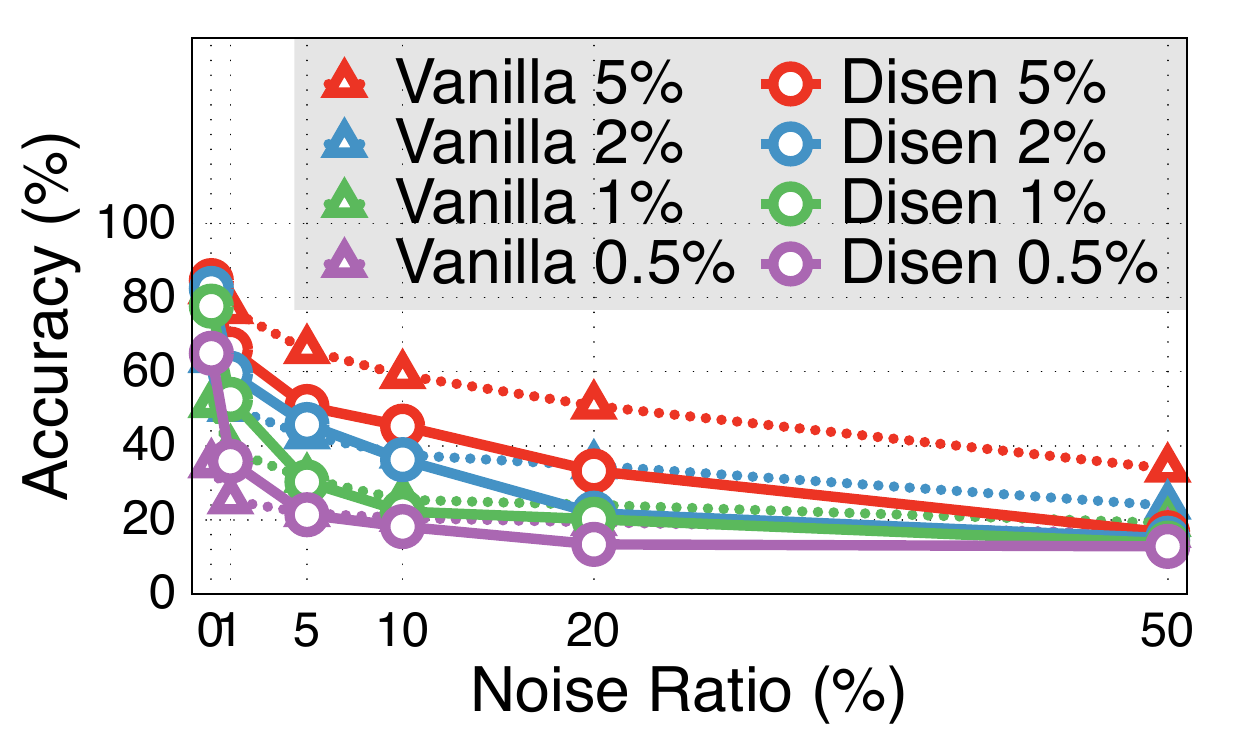}}
        \subfloat[\code{Entropy}]{\label{fig:entropy}\includegraphics[width=0.49\linewidth]{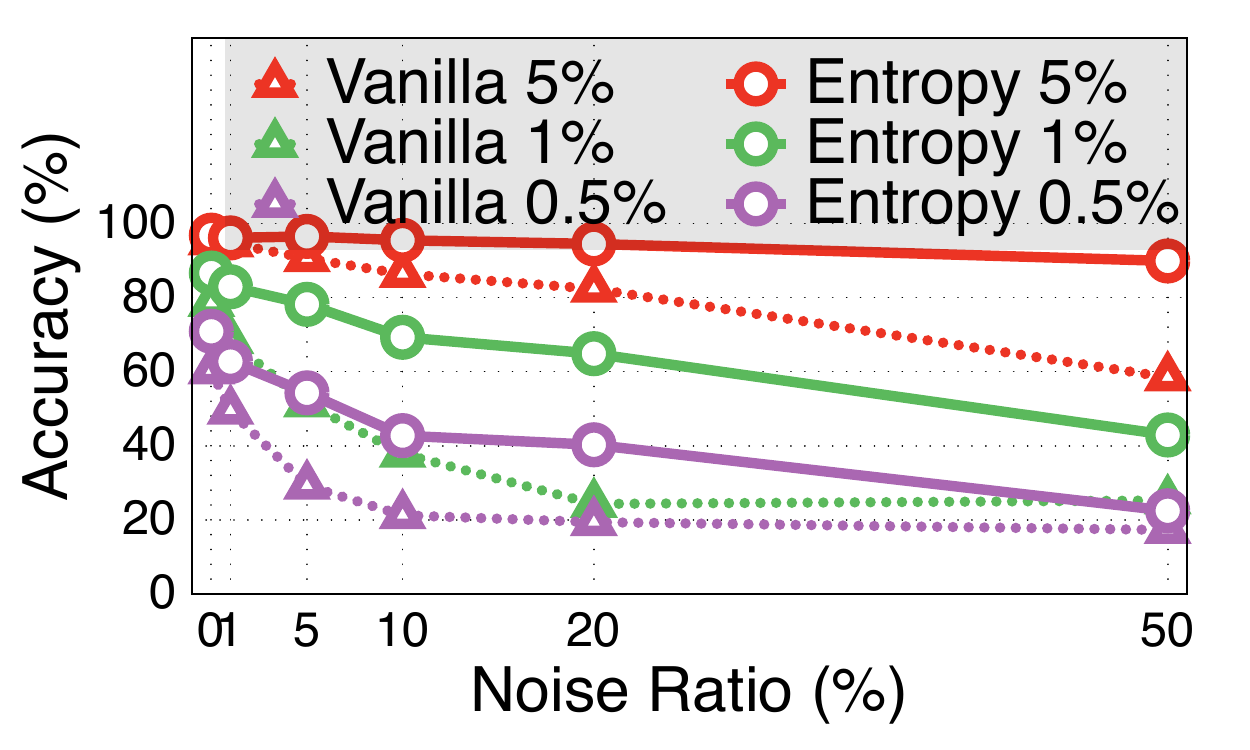}}
        \caption{Performance when label corruption occurs. In the case of \code{LfF} and \code{Disen}, it is the unbiased test accuracy of colored MNIST, and \code{JTT} is the worst case test performance of the waterbird dataset. The triangle-dotted lines are the vanilla results, and the circle-solid lines represent the result of each algorithm. All algorithms except for entropy case perform worse than vanilla as noise ratio $\eta$ increases.}
        \label{fig:degradation}
\end{figure}

\myparagraph{Brief summary of the previous methods.}
In previous debiasing algorithms, bias-conflicting samples are highlighted based on each proposed score. Almost all previous approaches train a biased model $f_b$ on the given training dataset, and a debiased model $f_d$ is trained with emphasis in the following ways:

\begin{itemize}[leftmargin=0.2in]
    \item Relative difficulty score (\code{LfF}~\cite{nam2020learning}, \code{Disen}~\cite{lee2021learning})
    \begin{equation}
        \label{eq:relative_diff}
        \mc{W}(x, \underline{\bm{y_\text{given}}}) = \frac{\mc{L}_\text{CE}(f_b(x),\underline{\bm{y_\text{given}}})}{\mc{L}_\text{CE}(f_b(x),\underline{\bm{y_\text{given}}}) + \mc{L}_\text{CE}(f_d(x),\underline{\bm{y_\text{given}}})},
    \end{equation}
    where $\mc{L}_\text{CE}$ denotes conventional cross-entropy loss and $f_b(\cdot)$ and $f_d(\cdot)$ are softmax outputs of biased and debiased models, respectively.
    \item Per-sample accuracy (\code{JTT}~\cite{liu2021just})
    \begin{equation}
        \label{eq:jtt_errorset}
        \mc{D}_\text{error-set} = \{(x, y) \text{ s.t. } \underline{\bm{y_\text{given}}} \neq  \argmax_c f_b(x)[c] \},
    \end{equation}
    where $f_b(x)[c]$ denotes the softmax output of logit $c$. The ultimate debiased model is trained on $\mc{D}_\text{train}$ composed of $\lambda_\text{up}$ times $\mc{D}_\text{error-set}$ and the other $\mc{D}_{\text{corr-set}} = \mc{D} \setminus \mc{D}_{\text{error-set}}$. 
\end{itemize}

\subsection{Debiasing meets noisy labels}

As in (\ref{eq:relative_diff}) and (\ref{eq:jtt_errorset}), all previous techniques are based on the given label $y_{\text{given}}$. Here, we refer to methods using $(x,y_{\text{given}})$ and only $(x)$ respectively as ``\emph{label-based debiasing}'' and ``\emph{label-free debiasing}.'' We observe the ultimate performance of previous methods when noisy labels are injected. We used the settings offered by their official repositories\footnote{https://github.com/alinlab/LfF}\footnote{https://github.com/anniesch/jtt}\footnote{https://github.com/kakaoenterprise/Learning-Debiased-Disentangled}, such as dataset, implementation, and hyperparameters except for label flipping. For comparison, we include entropy-based debiasing, which highlights samples with proportion to the per-sample entropy score. It does not require the given label, \ie label-free method. Detail description about entropy-based setting is described in Appendix.

\myparagraph{Label-based debiasing is prone to noisy labels.} As shown in Figure~\ref{fig:degradation}, the performances of the label-based techniques are lower than those of the vanilla case; a small noise ratio occurs. However, as demonstrated in Figure~\ref{fig:entropy}, the label-free method performs better than in the vanilla case. This is because label-based methods make incorrect emphasis, $\mc{W}(x,y_{\text{given}})$ and $\mc{D}_{\text{error-set}}$, when $y_{\text{given}}$ is corrupted. 


\begin{figure}[t!]
    \begin{minipage}{0.5\textwidth}
        \centering
        \includegraphics[width=0.6\textwidth]{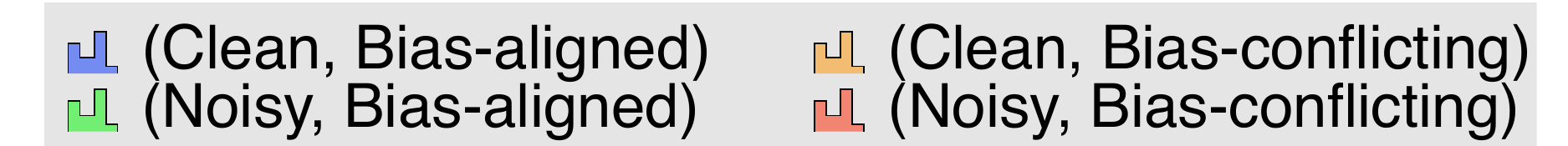}
    \end{minipage}
    \centering
    \subfloat[\code{LfF}~\cite{nam2020learning}]{\label{fig:lff_h}\includegraphics[width=0.49\linewidth]{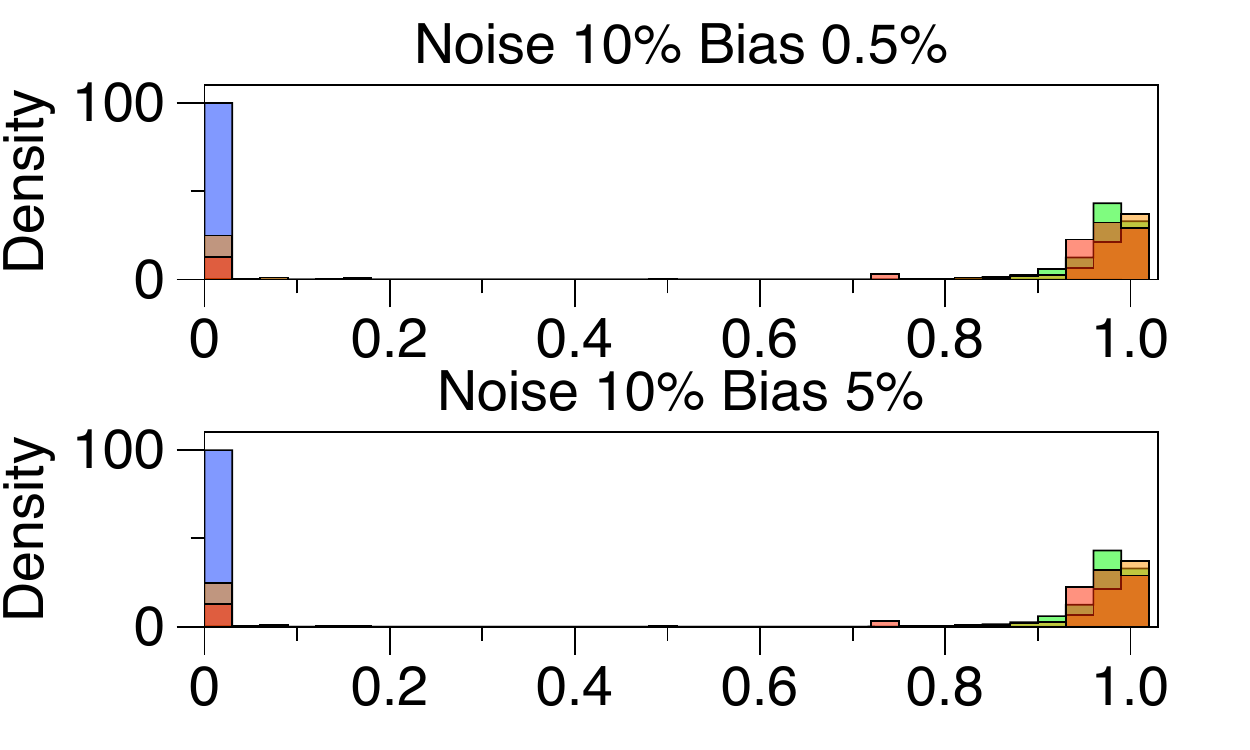}}
    \subfloat[\code{JTT}~\cite{liu2021just}]{\label{fig:jtt_h}\includegraphics[width=0.49\linewidth]{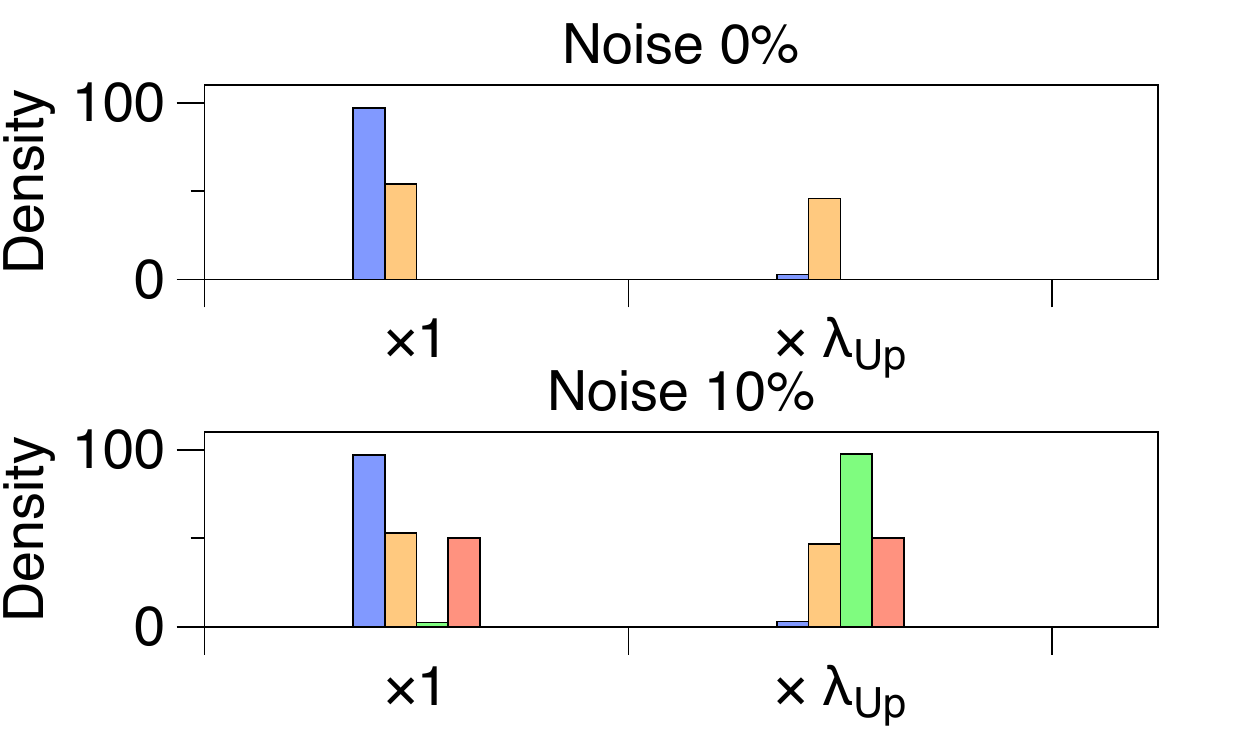}}\\
    \subfloat[\code{Disen}~\cite{lee2021learning}]{\label{fig:disen_h}\includegraphics[width=0.49\linewidth]{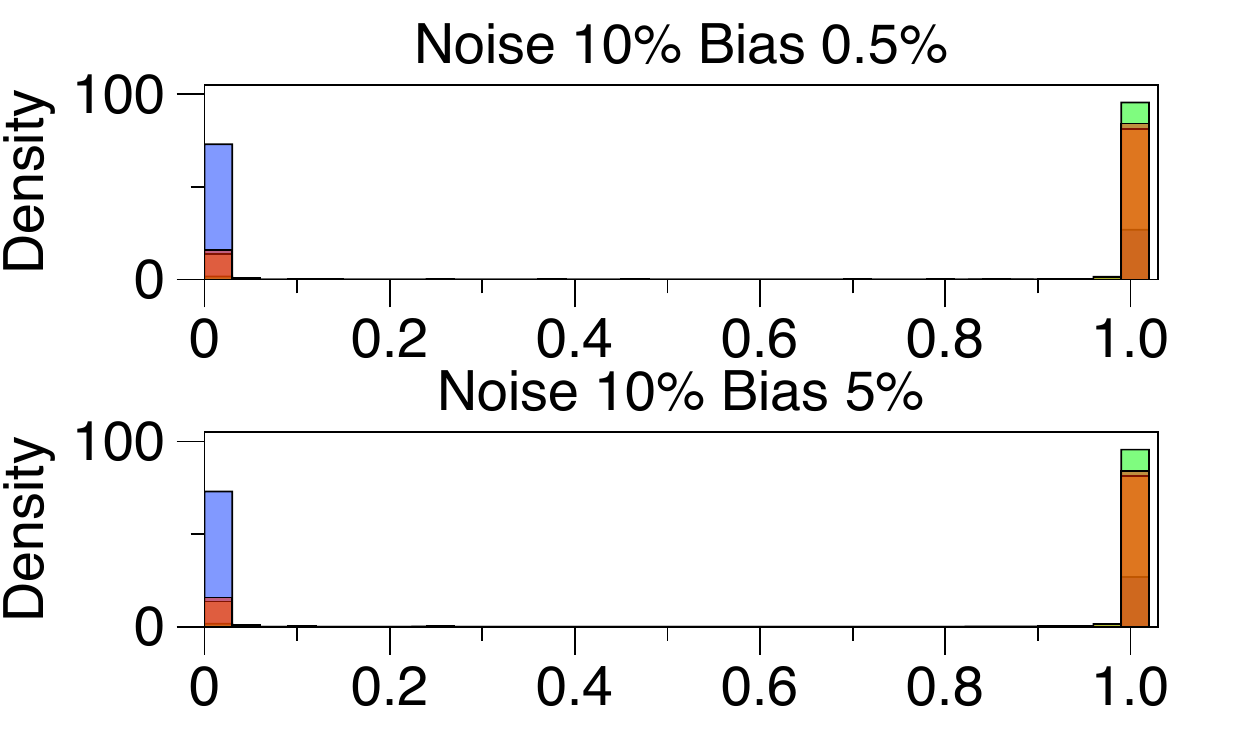}}
    \subfloat[\code{Entropy}]{\label{fig:entropy_h}\includegraphics[width=0.49\linewidth]{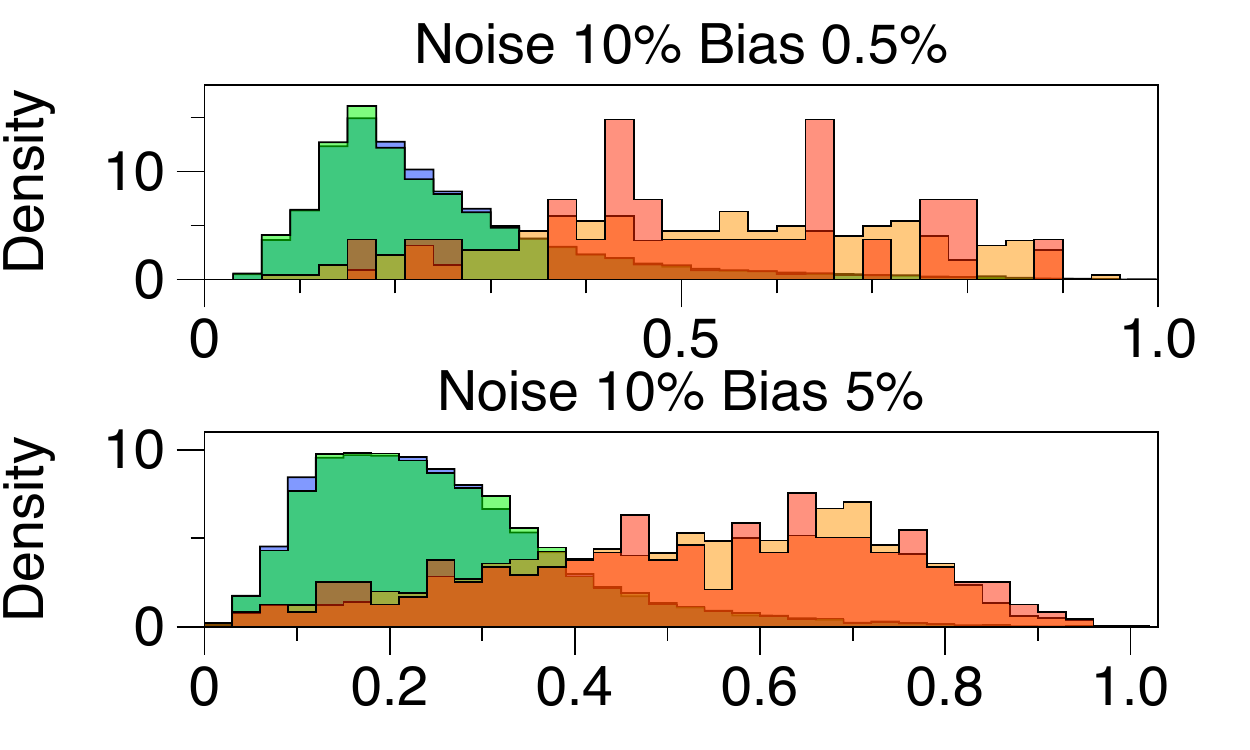}}
    \caption{Score histogram of each methodology. As \code{LfF} and \code{Disen} operate online, we report the weight histogram right after the last epoch of training. Except for the Entropy case, \code{LfF}, \code{JTT}, and \code{Disen} shows entangled histograms between \textcolor{green}{(noisy, aligned)}, \textcolor{orange}{(clean, conflicting)}, and  \textcolor{red}{(noisy, conflicting)}. By contrast, the histogram of entropy case indicates that it is clustered not according to label corruption but bias.}
    \label{fig:histogram}
\end{figure}

\myparagraph{Why do label-based methods suffer side-effects?}
As in Figure~\ref{fig:lff_h}~\ref{fig:jtt_h}, and~\ref{fig:disen_h}, the label-based scores of the \textcolor{orange}{(conflicting, clean)} and \textcolor{green}{(aligned, noisy)} are entangled. This means that noisy labels are also emphasized when we run the labe-based algorithms. By contrast, the label-free method, \ie \code{Entropy} in Figure~\ref{fig:entropy_h}, shows that the bias-conflicting and bias-aligned samples are easily distinguished regardless of whether their labels are noisy. In conclusion, a label-free method is needed to handle DBwNL.

%% file: main/denoise.tex
\section{\underline{DEN}oising after \underline{E}ntropy-based De\underline{B}iasing}
\label{sec:denoising}

In Section~\ref{sec:debiasing}, we verified that the debiasing algorithms do not work properly for DBwNL alone. In this section, we analyze how debiasing algorithms should be combined with denoising algorithms, and finally propose~\alg. 


\subsection{How denoising algorithms work in the DBwNL}
We first check how the denoising algorithms work in the DBwNL dataset by observing two cases. 

\myparagraph{Observation Setting.}
Five denoising methods are used for the analysis: \code{AUM}~\cite{pleiss2020identifying}, \code{Co-teaching}~\cite{han2018co}, \code{DivideMix}~\cite{li2019dividemix}, and \code{f-DivideMix}~\cite{kim2021fine}. We measure the number of samples of \textcolor{green}{(noisy, aligned)} and \textcolor{orange}{(clean, conflicting)} after running the denoising algorithms. We run two types of tests. (1) Without modification (\xmark): to check how the denoising algorithms handle bias-conflicting samples. (2) Manually weighted training ($\CIRCLE$): we assign $\times 50$ weights to bias-conflicting samples, \ie $50\times \mc{L}_{\text{CE}}(x,y)$, $\forall (x,y) \in \mc{D}_{c}$. The second case is unrealistic as we cannot know which sample is bias-conflicting, but the result can convey the following argument: if we want to protect bias-conflicting samples from the discarding by the denoising mechanism, make the bias-conflicting samples easy-to-learn.


\begin{figure}[t!]
    \begin{minipage}{0.5\textwidth}
        \centering
        \includegraphics[width=0.9\textwidth]{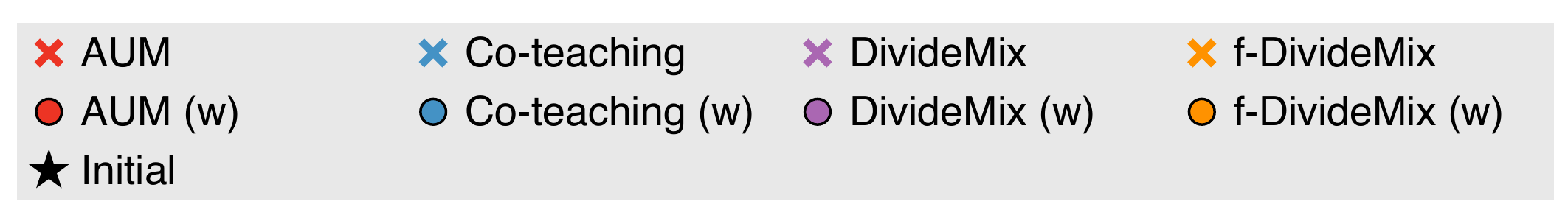}
    \end{minipage}
    \centering
    \subfloat[$\alpha = 1\%$, $\eta = 10\%$]
    {\label{fig:denoise_1_10}\includegraphics[width=0.49\linewidth]{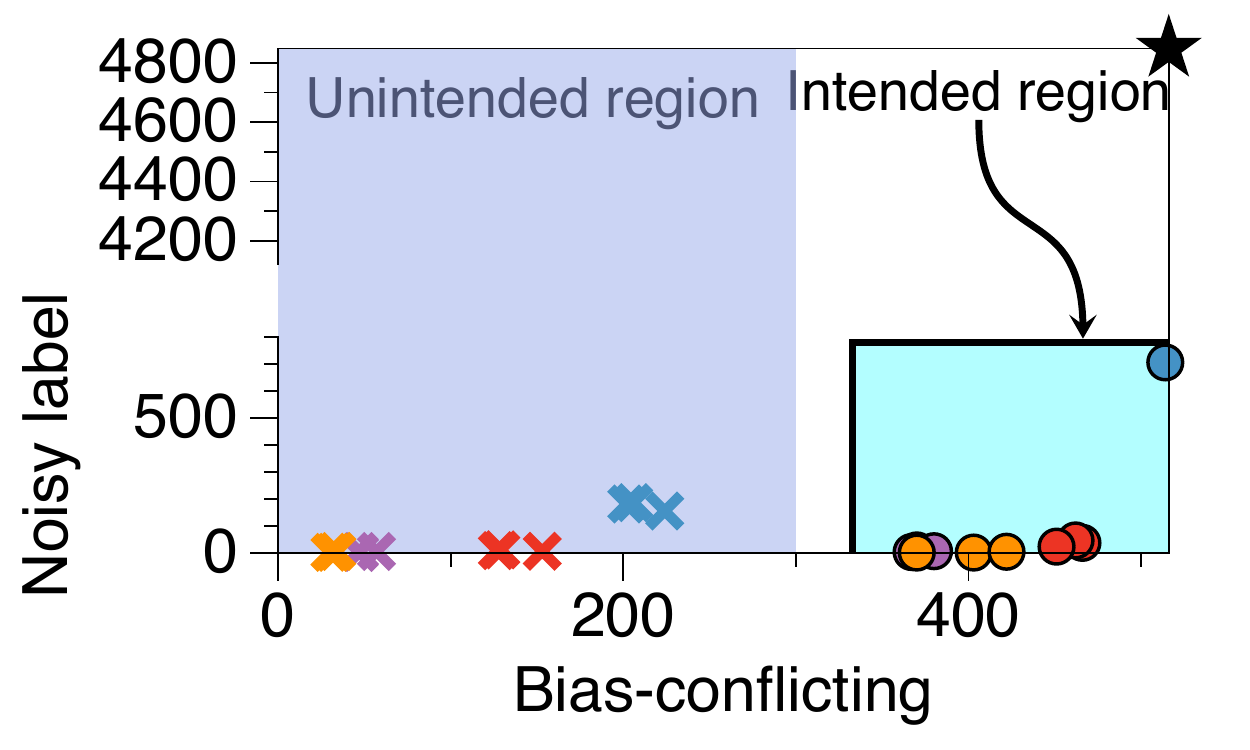}}
    \subfloat[$\alpha = 5\%$, $\eta = 50\%$]
    {\label{fig:denoise_5_50}\includegraphics[width=0.49\linewidth]{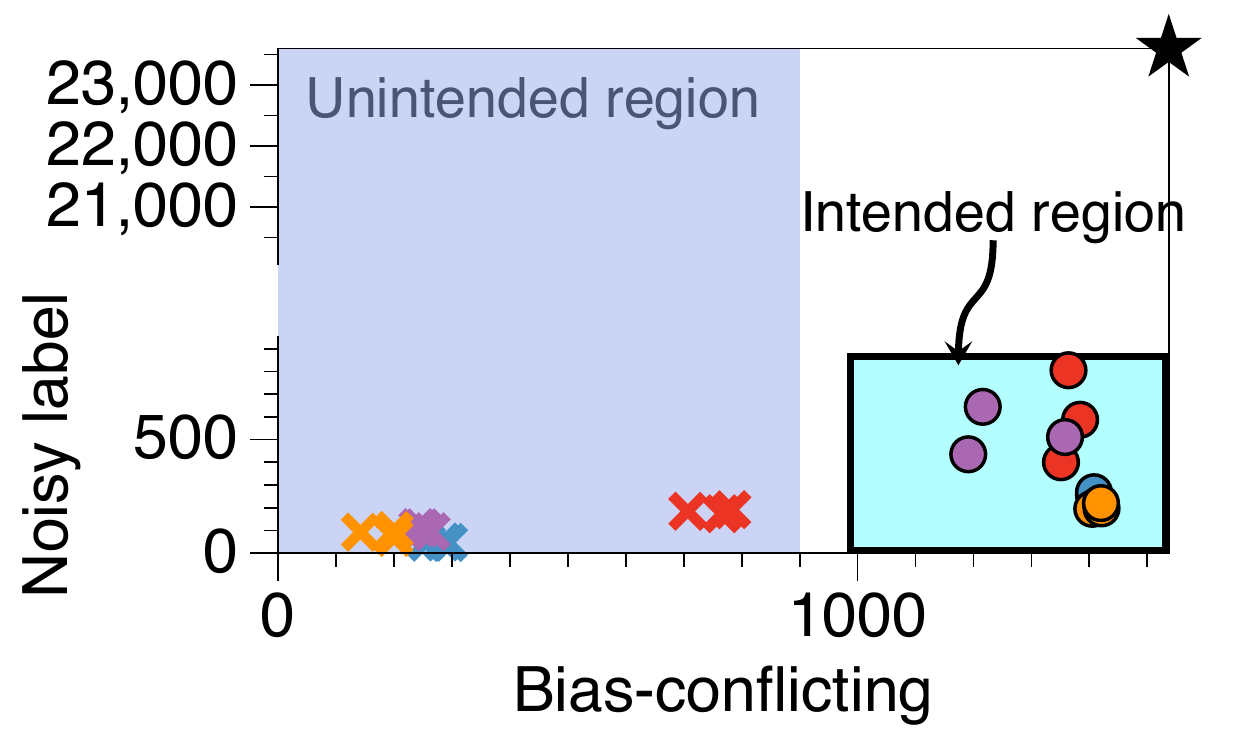}}
    \caption{Number of remaining noisy labels and bias-conflicting samples after denoising is conducted. Star $\star$ mark represents the number of samples before cleansing, and \xmark~ and $\CIRCLE$ marks indicate with or without weighted training results. Since bias-conflicting samples is precious for debiasing, bias-conflicting samples have to be protected. Therefore, the region loses bias-conflicting samples (left, blue) is the unintended region. On the other hand, the region ignores noisy labels without losing the bias-conflicting samples (right, cyan) is the intended behavior.}
    \label{fig:denoise}
\end{figure}

\myparagraph{Valuable bias-conflicting samples can be deemed noisy.}
As illustrated in Figure~\ref{fig:denoise}, all denoising methods sufficiently differentiate noisy samples. For example, $\star$ in Figure~\ref{fig:denoise_1_10} represents that the initial number of noisy samples is almost $5,000$, but almost all methods dropped to near $0$ after denoising (\xmark~ marks). However, crucial bias-conflicting samples are also eliminated, \ie \xmark~ marks are in the ``unintended region.'' Therefore, utilizing denoising algorithms before debiasing can discard valuable bias-conflicting samples. As in Figure~\ref{fig:degradation}, because the number of bias-conflicting samples is critical, removing the bias-conflicting samples prior to debiasing can cause performance degradation.

\myparagraph{Preventing bias-conflicting samples from being discarded.}  
Bias-conflicting samples are considered noisy labels because the trained model thinks that they are difficult-to-learn. As illustrated in Figure~\ref{fig:denoise} using $\CIRCLE$ marks, when we sufficiently highlight bias-conflicting samples, noisy labels can be eliminated by reducing the loss of bias-conflicting samples. Therefore, we can conclude that denoising should be performed after highlighting bias-conflicting samples.

%% file: main/method.tex

\subsection{Designing the Algorithm for DBwNL}

\begin{figure}[t]
        \centering
        \includegraphics[width=1.0\columnwidth]{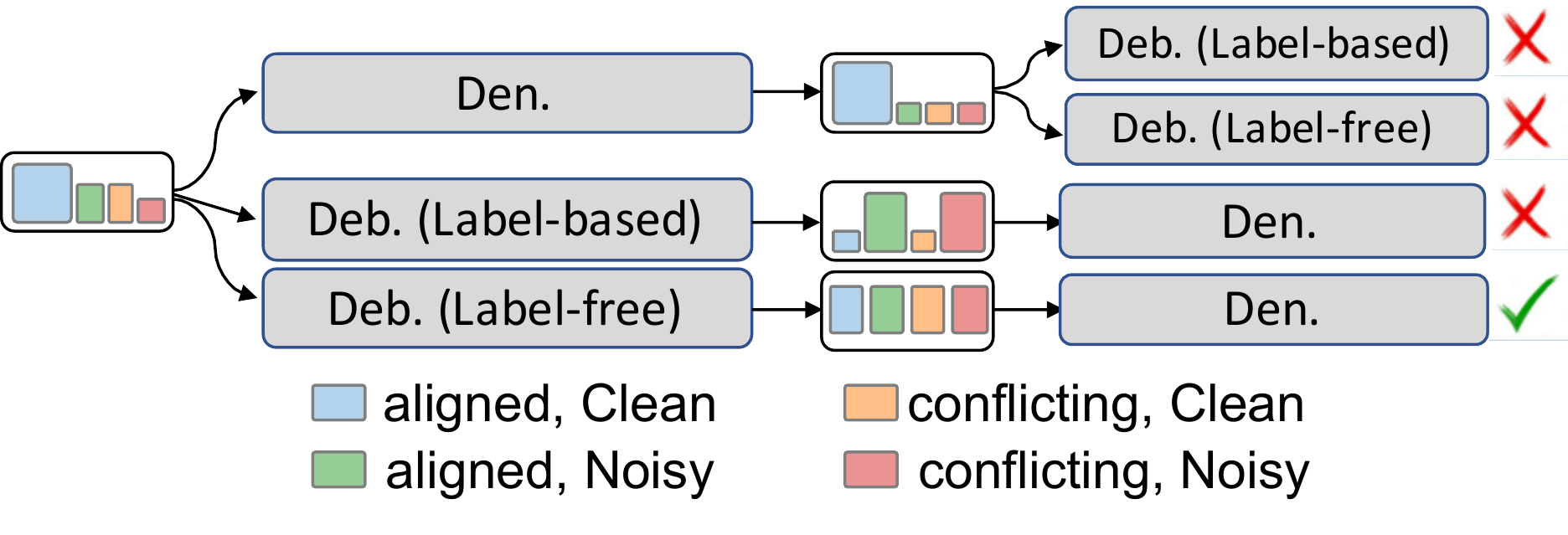}
        \caption{Case study of designing algorithm for DBwNL. }
        \label{fig:intuition}
\end{figure}

Based on the previous experimental results, two inferences can be drawn in designing a debiasing algorithm for the DBwNL datasets. (1) No label-based: label-based debiasing emphasizes noisy labels. (2) Debiasing before denoising: denoising algorithms should be run after debiasing emphasizes bias-conflicting samples. We summarize our intuitions in Figure~\ref{fig:intuition}. For the DBwNL dataset, the main consideration is whether to apply debiasing or denoising first. If denoising is applied first, the bias-conflicting samples is erased, which is burdensome for debiasing (see the lower $\alpha$ cases in Figure~\ref{fig:degradation}). Conversely, if debiasing is conducted first, we can choose label-based or label-free. If a label-based algorithm is selected, the noisy labels are enlarged and a burden is placed on the denoising algorithm (see the higher $\eta$ cases in Figure~\ref{fig:degradation}). In other words, emphasizing the bias-conflicting sample and proceeding with denoising without emphasizing the noisy label through the label-free algorithm is the correct order.


\subsection{Denoising after entropy-based Debiasing}
Based on the case study, we propose denoising after entropy-based debiasing, \alg, which is composed of three steps.

\myparagraph{Step 1: Train the prejudice model $f_p$.}
The key aim while training the prejudice model $f_p$ is that regardless of label corruption, the model should comprehensively learn the bias-aligned samples so that it can identify the bias-conflicting samples in the next steps. However, it is difficult to detach \textcolor{green}{(bias-aligned, noisy)} from the bias-conflicting samples. Therefore, \alg~ trains the prejudice model on \textcolor{blue}{(bias-aligned, clean)} only. Intuitively speaking, if the model is trained using only \textcolor{blue}{(bias-aligned,clean)} samples, \textcolor{green}{(bias-aligned,noisy)} samples also can be regarded as easy-to-learn thanks to the bias attributes. \alg~ finds \textcolor{blue}{(bias-aligned, clean)} by using the GMM, similar with~\cite{li2019dividemix, kim2021fine}. Step 1 consists of two sub-steps. At first, $f_p$ is trained on $\mc{D}$ with conventional cross-entropy loss, until the warm-up epoch $e_w$. After the warm-up phase, \alg~ splits $\mc{D}$ and obtains $\bar{\mc{D}}$ at the beginning of each epoch. To do so, \alg~ dynamically fits a GMM on per-sample losses and obtains $\bar{\mc{D}}$ whose probability of GMM $g(x_i,y_i)$ is higher than the threshold $p_t$ at the beginning of each epoch:
\begin{equation}
    \bar{\mc{D}} = \{(x_i,y_i) | g(x_i,y_i) > p_t, \text{ where } (x_i,y_i) \in \mc{D} \},
\end{equation}
Note that, unlike \code{DivideMix} and \code{f-DivideMix}, \alg~does not use the samples whose $g(x_i,y_i) \leq p_t$, to deepen bias, \ie it ignores every-types except for \textcolor{blue}{(bias-aligned,clean)}.

\myparagraph{Step 2: Calculate sampling probability.}
Based on the trained prejudice model $f_p$, we extract the entropy score for each sample:
\begin{equation}
    \label{eq:weight}
        H_{\tau}(x)  = -\sum_{c}^C f_p(x,\tau)[c] \times \log f_p(x,\tau)[c], 
\end{equation}
where $f_p(x,\tau)[j]$ is the temperature-scaled softmax for class $c$ with temperature parameter $\tau$, \ie $f_p(x,\tau)[j] = \frac{\exp (q_p(x)[j]/\tau)}{\sum_c \exp (q_p(x)[c]/\tau)}$ with logit $q_p(x)$. Based on $H_{\tau}(x)$, we find the sampling probability of each instance as follows:
\begin{equation}
    \label{eq:samprob}
    \mc{P}(x_i,y_i) = \frac{H_{\tau}(x_i)}{\sum_{(x_j,y_j) \in \mc{D}} H_{\tau}(x_j)}.
\end{equation}
The reason why $\mc{P}(x_i,y_i)$ is proportional to the entropy score is because $f_p$ is sufficiently biased and thus the larger entropy samples are the bias-conflicting samples (see Figure~\ref{fig:entropy_h}).

\myparagraph{Step 3: Train the robust model $f_r$.} 
To train the robust model $f_r$, mini-batches are constructed based on the sampling probability in~\eqref{eq:samprob}. As mini-batches contain sufficient bias-conflicting samples, the main purpose of this step is to mitigate the impact of noisy labels. To this end, we inherit previous denoising algorithms by simply modifying the mini-batches. Note that \alg~ can utilize any given denoising algorithm, $\mc{A}_{\text{den}}$, but we report based on \code{GCE} which performs better than the others. As analysis of various denoising algorithms is reported in Section~\ref{sec:exp}.

%% file: main/exp.tex
\section{Experiment}
\label{sec:exp}

The effectiveness of the proposed algorithm is analyzed quantitatively and qualitatively. We compare \alg~to earlier debiasing and denoising techniques in four biased datasets, \ie Colored-MNIST (CMNIST), Corrupted CIFAR-10 (CCIFAR), Biased Action Recognition (BAR), and Biased FFHQ (BFFHQ). To test the generalization performance, we report the unbiased test accuracy. Details of the implementation and datasets are given in Appendix.

\input{table/benchmark_summary}

\input{table/noise_synth_alone}
\input{table/real}

\subsection{Experimental settings}

\myparagraph{Baselines.} We report the performance of the debiasing and denoising algorithms. As debiasing algorithms, we use recent methods that are officially available, such as~\code{LfF}, \code{JTT}\footnote{As~\cite{liu2021just} assume that a balanced validation dataset, it is unfair to directly compare with the other algorithms. However, since \code{JTT} can be tuned using noisy biased validation dataset, we report the behavior of \code{JTT} tuned by using a biased noisy vallidation.}, \code{EIIL}, and \code{Disen}.
We utilize denoising algorithms~\code{GCE}, \code{SCE}, \code{ELR+}, \code{Co-teaching}, \code{DivideMix}, and \code{f-DivideMix}. All implementations are reproduced following the official codes. The implementation and hyperparameters are reported in Appendix.

\myparagraph{Datasets.} We employ four benchmarks: CMNIST, CCIFAR, BAR, and BFFHQ. The target and bias attributes are summarized in Table~\ref{tab:benchmark}. For the CMNIST and CCIFAR datasets, two pairs of bias-ratio $(\alpha,\eta) = \{ (1\%,10\%), (5\%,50\%)\}$ are utilized. The other datasets are tested on $\eta=10\%$. We summarize in detail the construction recipe in Appendix.

\myparagraph{CMNIST and CCIFAR.} CMNIST and CCIFAR datasets have a bias attribute, which is injected manually. The goal of CMNIST is to classify the target attribute, \code{digit shape}, when the bias attribute is \code{color}. This dataset comes from~\cite{nam2020learning, bahng2020learning, lee2021learning, kim2021biaswap}. In CCIFAR, the target attribute is \code{objective} such as \{\code{airplane}, \code{car},...\} with the bias attribute \code{corruption} like \{\code{blur}, \code{gaussian noise}, ...\}. We generate CCIFAR following the bias injection mechanism of~\cite{nam2020learning, lee2021learning, hendrycks2018benchmarking}.

\myparagraph{BAR and BFFHQ.} BAR and BFFHQ are consists of real-world images. These benchmarks are biased when selecting samples by seeing the multiple attributes. BAR~\cite{nam2020learning} aims to classify actions such as \{\code{racing}, \code{climbing},...\} with background bias. For example, \code{(Climbing, Rockwall)} are bias-aligned samples, while \code{(Climbing, Ice-cliff)} are the bias-conflicting ones. BFFHQ~\cite{kim2021biaswap, lee2021learning} aims to classify gender when its age is biased. For example, the training dataset is made up of (\code{Female, Young} (age ranging from 10 to 29)) and (\code{Male, Old} (age ranging from 40 to 59)) and very few of \code{(Female, Old)} and \code{(Male, Young)} samples. 

\myparagraph{Implementation details.}
For the Colored MNIST, we use a Simple-ConvNet with three convolutional layers, ReLU activation function~\cite{agarap2018deep}, batch normalization~\cite{ioffe2015batch} and dropout~\cite{srivastava2014dropout}. ResNet-18~\cite{he2016deep} pre-trained on ImageNet is used as a backbone network for the rest. Based on grid searches, we find hyperparameters for all algorithms using $90\%$ and $10\%$ training and validation split. This implies that validation datasets contain noisy labels and bias-conflicting samples. The search space and searched hyperparameters are described in Appendix. For all experiments, we report the case where \alg~ uses \code{GCE} as $\mc{A}_{\text{den}}$, which achieves the best performance among all the denoising algorithms.

\subsection{Experimental results}

\myparagraph{CMNIST and CCIFAR.}
Table~\ref{tab:noise_synth} presents comparisons of the accuracy of the unbiased test. Among the debiasing baselines, the accuracy-based algorithm, \ie \code{JTT}, is better than the \code{vanilla} model in the CMNIST case. However, all debiasing baselines obtain worse performance than the \code{vanilla} model in the CCIFAR case because, as mentioned earlier, the debiasing algorithms highlight noisy labels that should not be emphasized. By contrast, denoising algorithms fail to debias, as they do not have a module to highlight bias-conflicting samples. \alg~ achieves the best performance for all injected bias cases. For example, the unbiased test accuracy of CMNIST with $\alpha=1\%$ and $\eta=10\%$ shows that \alg~ obtains $52.57\%$ gain compared to the \code{Vanilla} model.

\myparagraph{BAR and BFFHQ.}
The performances of \alg~ in real-world image benchmarks is also better than the other baselines. BAR shows $7.92\%$ improvements over vanilla and $6.5\%$ improvement over \code{Disen}, which has the best performance among the debiasing algorithms. \alg~  also shows $5.91\%$ performance gain over \code{GCE}. Similarly, BFFHQ shows $3.86\%$ improvement over vanilla, $5.06\%$ over \code{JTT} and $3.04\%$ over \code{DivideMix}. Thus, entropy-based debiases when trained on a more complex raw image dataset.

\begin{figure}[h!]
    \centering
    \includegraphics[width=\columnwidth]{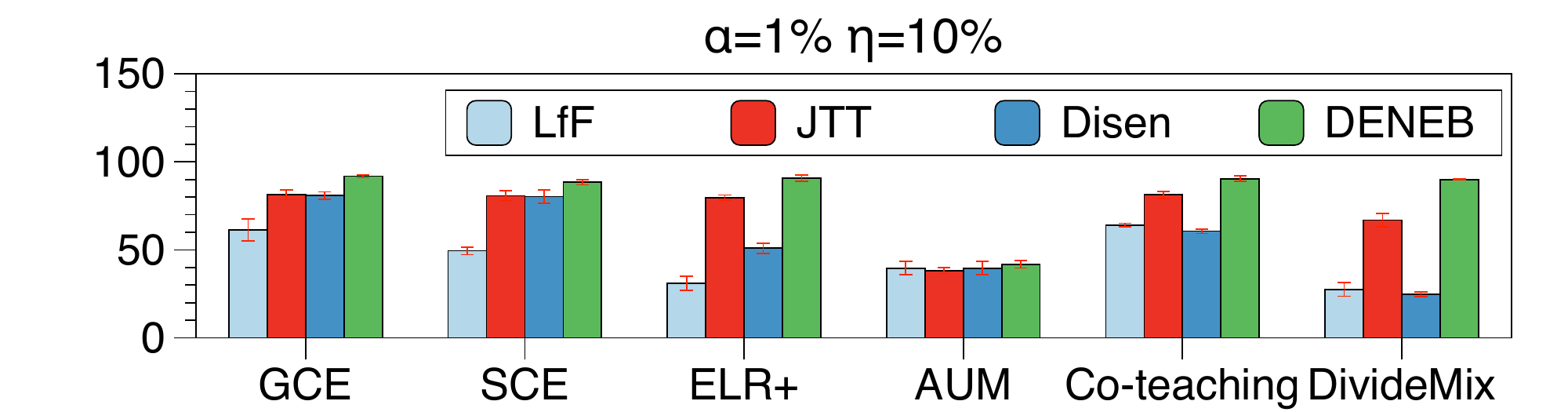}\\
    \includegraphics[width=\columnwidth]{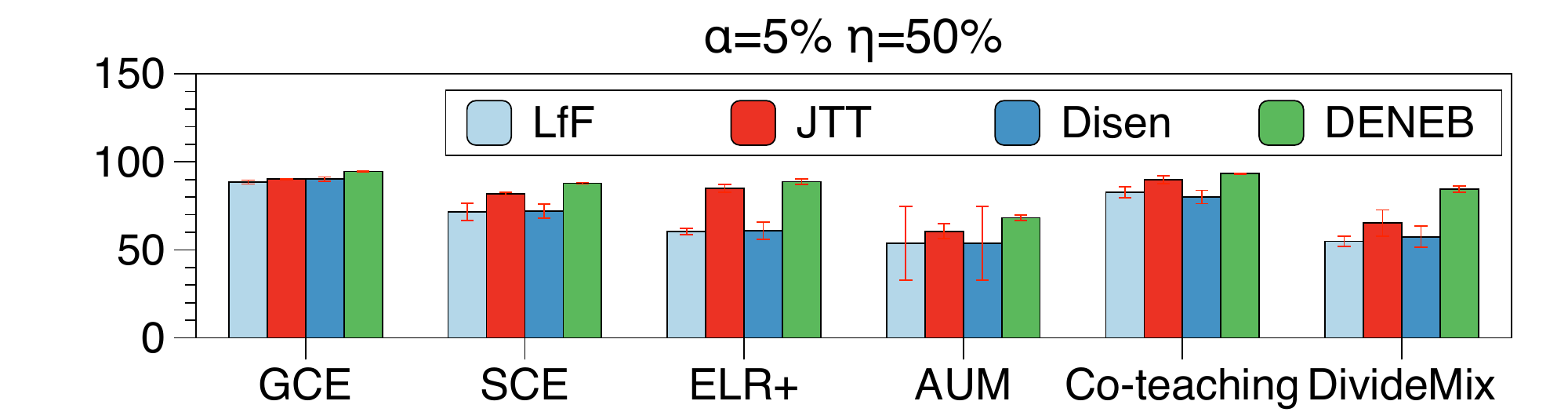}
    \caption{Combination result of Colored MNIST benchmark. All cases are the performances of Debiasing $\to$ Denoising, \ie obtain per-sample weights from \alg and then run \code{GCE} for \alg$\to$\code{GCE} case.}
    \label{fig:combination}
\end{figure}

\myparagraph{Combination of Debiasing and Denoising.} In order to study how the other debiasing algorithms work with denoising algorithms, \ie debiasing $\to$ denoising similarly to \alg, we report pairwise performance in Figure~\ref{fig:combination} for CMNIST. As \code{Disen} and \code{LfF} are an online algorithms, we multiply the per-sample weight at the end of debiasing by denoising loss. Details are provided in Appendix. As shown in Figure~\ref{fig:combination}, \alg~ performs better than the other debiasing algorithms for all combinations. \alg~ has better performance because the side-effects of focusing on the noisy sample are minimized when using the label-free entropy.


%% file: table/benchmark_summary.tex
\begin{table}[!h]
    \caption{Benchmark Summary}
    \label{tab:benchmark}
    \centering
    \small{
        \begin{tabular}{ccccccc}
        \thickhline
        Dataset & train/valid/test & \#class & Target & Bias  \\ \hline
        CMNIST & 54K / 6K / 10K & 10 & Shape & Color   \\ 
        CCIFAR & 45K / 5K / 10K & 10 & Object & Blur   \\
        BAR & 1,746 / 195 / 654 &  6 & Action & Place \\
        BFFHQ & 17,280/1,920/1,000 & 2 & Gender & Age \\
        \thickhline
        \end{tabular}
    }
\end{table}

%% file: table/noise_synth_alone.tex
\begin{table*}[!tb]
    \caption{Unbiased test accuracy on CMNIST and CCIFAR. Best-performing results are marked in bold. All results are averaged on three independent runs. \alg~represents \ie $\mc{A}_{\text{den}} = $ \code{GCE}.}
    \label{tab:noise_synth}
    \centering
    \small{
    \begin{tabular}{ccccc}
        \thickhline
        \multirow{2}{*}{\bfcode{Algorithm}} & \multicolumn{2}{c}{Colored MNIST}     & \multicolumn{2}{c}{Corrupted CIFAR-10} \\
        & $\alpha = 1$\%, $\eta = 10\%$    & $\alpha = 5$\%, $\eta = 50\%$                    
        & $\alpha = 1$\%, $\eta = 10\%$    & $\alpha = 5$\%, $\eta = 50\%$ \\ \hline
        \bfcode{Vanilla}             &39.24 {   $\pm$ 1.91}\%             &70.13\% {   $\pm$ 3.42} \%  &25.43\% {   $\pm$ 0.84} \%      &31.86\% {   $\pm$ 0.96} \% \\  \hline
        \multicolumn{5}{c}{\bfcode{Debiasing}} \\ \hline
        \bfcode{LfF}             &29.87 {   $\pm$ 1.36}\%          &57.97\% {   $\pm$ 1.79} \%  &24.51\% {   $\pm$ 1.30} \%            &29.68\% {   $\pm$ 2.63} \%\\ 
        \bfcode{JTT}         &63.24 {   $\pm$ 2.60}\%             &77.16\% {   $\pm$ 1.15} \%  &23.75\% {   $\pm$ 0.61} \%       &24.52\% {   $\pm$ 0.98} \%\\ 
        \bfcode{EIIL}         &24.53 {   $\pm$ 0.31}\%             &42.25\% {   $\pm$ 1.43} \%  &20.30\% {   $\pm$ 1.08} \%       &22.66\% {   $\pm$ 1.94} \%\\ 
        \bfcode{Disen}       &31.49 {   $\pm$ 5.44}\%          &69.20\% {   $\pm$ 4.13} \%  &22.52\% {   $\pm$ 0.38} \%       &28.35\% {   $\pm$ 4.49} \%\\ 
                                        \hline
        \multicolumn{5}{c}{\bfcode{Denoising}} \\ \hline
        \bfcode{GCE}        &19.52 {   $\pm$ 1.98}\%       &73.45\% {   $\pm$ 7.62} \%  &24.96\% {   $\pm$ 1.53} \%          &30.72\% {   $\pm$ 0.74} \%\\ 
        \bfcode{SCE}        &30.95 {   $\pm$ 2.87}\%            &62.10\% {   $\pm$ 5.02} \%  &23.34\% {   $\pm$ 1.73} \%     &29.87\% {   $\pm$ 1.00} \%\\ 
        \bfcode{ELR+}       &24.76 {   $\pm$ 0.90}\%             &49.38\% {   $\pm$ 3.74} \%  &22.10\% {   $\pm$ 0.37} \%           &30.84\% {   $\pm$ 0.43} \%\\ 
        \bfcode{AUM}&23.89 {   $\pm$ 2.60}\%              &49.51\% {   $\pm$ 6.62} \%  &23.55\% {   $\pm$ 1.10} \%              &28.06\% {   $\pm$ 2.38} \%\\ 
        \bfcode{Co-teaching}&41.89 {   $\pm$ 1.45}\%              &76.64\% {   $\pm$ 5.52} \%  &25.14\% {   $\pm$ 0.27} \%              &26.84\% {   $\pm$ 0.52} \%\\ 
        \bfcode{DivideMix}&20.48 {   $\pm$ 1.94}\%        &33.66\% {   $\pm$ 2.91} \%  &18.86\% {   $\pm$ 0.28} \%           &22.03\% {   $\pm$ 0.59} \%\\ 
        \bfcode{f-DivideMix}&22.06 {   $\pm$ 1.70}\%             &39.92\% {   $\pm$ 3.26} \%  &19.67\% {   $\pm$ 0.25} \%              &27.60\% {   $\pm$ 0.54} \%\\ 
                                        \hline
        \multicolumn{5}{c}{\bfcode{\alg}}\\\hline
        \bfcode{\alg}             &\textbf{91.81 {   $\pm$ 0.84}\%}       &\textbf{94.55\% {   $\pm$ 0.22} \%}  &\textbf{26.05\% {   $\pm$ 0.54} \%}                & \textbf{35.32\% {   $\pm$ 1.03} \%} \\ \hline
        \thickhline
        \end{tabular}
        }
\end{table*}

%% file: table/real.tex
\begin{table}[!tb]
    \caption{Unbiased test accuracy on BAR and BFFHQ. Best performing results are marked in bold. All results are averaged on three independent runs.}
    \label{tab:real}
    \centering
        \begin{tabular}{ccc}
        \thickhline
        \multirow{2}{*}{\bfcode{Algorithm}}  & BAR & BFFHQ \\ 
            & $\eta = 10\%$ & $\eta = 10\%$  \\ \hline
         \bfcode{Vanilla} &   54.37 { $\pm$ 1.10}\%     & 71.38 { $\pm$ 0.58}\%    \\ \hline
         \bfcode{LfF}  & 53.62 { $\pm$ 1.81}\%     & 54.35 { $\pm$ 0.91}\%    \\
         \bfcode{JTT}  & 55.67 { $\pm$ 2.16}\%     & 70.18 { $\pm$ 1.47}\%    \\
         \bfcode{Disen} &  55.80 { $\pm$ 3.05}\%     & 67.44 { $\pm$ 2.57}\%    \\ \hline
         \bfcode{GCE} &  56.39 { $\pm$ 0.95}\%     & 68.45 { $\pm$ 2.98}\%    \\
         \bfcode{Co-teaching} &  54.99 { $\pm$ 1.28}\%     & 69.28 { $\pm$ 1.24}\%    \\
         \bfcode{DivideMix} &  52.01 { $\pm$ 1.51}\%     & 72.20 { $\pm$ 0.58}\%    \\ \hline
         \bfcode{\alg} &  \textbf{62.30 { $\pm$ 0.91}\%}     & \textbf{75.24 { $\pm$ 0.68}\%}    \\
        \thickhline
        \end{tabular}
\end{table}


%% file: main/related.tex
\section{Related work}
\label{sec:related}


\myparagraph{Noisy labels.} \cite{ghosh2017robust} had proposed the mean absolute error (MAE), and~\cite{zhang2018generalized} claim that MAE suffers poor robustness with DNN and suggested another type of cross-entropy (CE) loss, called generalized cross-entropy (GCE). The authors of~\cite{wang2019symmetric} propose symmetry cross-entropy (SCE) loss, which is a combination of conventional CE and reverse cross-entropy. Lukasik et al.~\cite{lukasik2020does} use label smoothing techniques for noisy labels. Recently, studies on the early learning phase have been a topic of extensive interest. These works claim that DNNs memorize difficult-to-learn samples in the later phase and learn common features in the early learning phase.
Based on this fact,~\cite{liu2020early} propose the early learning regularizer (ELR) to prohibit memorizing noisy labels. Some works handle noisy labels by detecting and cleansing. To do so, the co-training method, \ie teaching each other, is mainly used. In~\cite{han2018co} and~\cite{yu2019does}  utilize loss and disagreement are utilized to construct a clean subset. 
\cite{pleiss2020identifying} proposes a new metric, area under margin (AUM), to cleanse the dataset. \cite{li2019dividemix} look at the noisy label problem as a semi-supervised learning (SSL) approach by dividing the training dataset into clean labeled and noisy unlabeled sets, and running the SSL algorithm~\cite{berthelot2019mixmatch}. FINE~\cite{kim2021fine} uses the alignment of the eigenvector to distinguish clean and noisy samples.


\myparagraph{Debiasing with human supervision.} \cite{goyal2017making, goyal2020rel3d} construct a debiased dataset with the human hand. \cite{alvi2018turning, kim2019learning, mcduff2019characterizing, singh2020don, li2018resound, li2019repair} use bias labels to mitigate the impact of bias labels when classifying target labels. EnD~\cite{tartaglione2021end} proposes to entangle the target attribute and disengle the biased attributes. Multi-expert approaches~\cite{alvi2018turning, kim2019learning, teney2021evading} use a shared feature extrator with multiple FC layers to classify multiple attributes independently. \cite{mcduff2019characterizing, ramaswamy2021fair} use conditional generator to determine if the trained classifier is biased. \cite{singh2020don} proposes overlap loss, which is measured based on the class activation map. 
\cite{li2019repair} employs bias type to detect bias-conflicting samples and reconstruct balanced dataset. On the other hand,~\cite{geirhos2018imagenet, wang2019learning, lee2019simple} using prior knowledge of the bias context to mitigate dataset bias. 
\cite{liu2021just, zhang2022correct} use bias labels for validation datasets to tune the hyperparameters.

\myparagraph{Debiasing without human supervision.} To reduce human intervention,~\cite{bras2020adversarial, kim2021biaswap, idrissi2022simple} utilize per-sample accuracy. They regard the inaccurate samples as bias-conflicting. \cite{lee2021learning, nam2020learning} use the loss to calculate the weight. In this case, samples with higher loss from the biased model are overweighted when training the debiased model. \cite{creager2021environment, sohoni2020no} infer the bias-conflicting labels and use the predicted labels to mitigate dataset bias problem.  \cite{darlow2020latent} generate the samples whose loss becomes large using variational auto-encoder. \cite{zhang2022rich} propose an initialization point for enlarging the features. 

%% file: main/conclusion.tex
\section{Conclusion}
\label{sec:conclusion}

Dataset bias with noisy labels can degrade prior debiasing algorithms. To overcome this issue, we propose \alg comprising three stpes. First, the prejudice model is trained on the clean bias-aligned samples. To do so, we utilize a GMM model to select clean bias-aligned samples. After training the prejudice model, \alg~ compute the entropy score for each sample. This entropy score does not require labels, which can mislead the algorithm into detecting bias-conflicting samples. Based on the obtained entropy score, we compute a sampling probability proportional to the entropy score. To train the final robust model, mini-batches are constructed with sampling probabilities and existing denoising algorithms are run based on the sampled mini-batches. Through extensive experiments across multiple datasets, such as Colored MNIST, Corrupted CIFAR-10, BAR, and BFFHQ, we show that \alg consistently obtains substantial performance improvements compared to the other algorithms for the debiasing, denoising, or na\"ively combined method. For future work, we plan to adapt this algorithm to other domains such as NLP, VQA, and so on. We hope that this study opens the door of training a robust model on DBwNL dataset.

%% file: appendix/appendix.tex
\appendix
\setcounter{footnote}{0} 
\setcounter{table}{0} 
\setcounter{figure}{0} 
\onecolumn

This is supplementary material for ``Denoising after Entropy-based Debiasing A Robust Training Method for Dataset Bias with Noisy Labels.'' This material presents additional results and descriptions that are not included in the main paper owing to the page limit. Section~\ref{app:pseudocode} describes the pseudocode of \alg. Section~\ref{app:exp} summarizes experimental settings. Section~\ref{app:obs_setting} shows experimental details for Section~\ref{sec:debiasing} and Section~\ref{sec:denoising}. Section~\ref{app:compute_resource} summarizes computation resources and training time for running \alg.

\input{appendix/pseudocode}
\input{appendix/exp_setting}
\input{appendix/obs_setting}
\input{appendix/computing}

%% file: appendix/pseudocode.tex
\section{Pseudo-code}
\label{app:pseudocode}

\begin{algorithm}[!h]
    \DontPrintSemicolon
    \SetAlgoLined
    \SetNoFillComment
    \LinesNotNumbered 
    \caption{Deneb: \underline{Den}oising after \underline{e}ntropy-based de\underline{b}iasing}
    \label{alg:deneb}
    \KwInput{Training set $\mc{D}$, Warmup epoch $e_w$, GMM threshold $p_t$, Denoising algorithm $\mc{A}_\text{den}$, Epoch $E$, GCE parameter $q$, Temperature parameter $\tau$.}
    \KwOutput{Debiased model.}
    
    \While{$e < E$\tcp*{Train prejudice model $f_p$}} 
    {   
        \If{$e < e_w$}
        {
            Upate $f_p$ using $\mc{L}_{\text{CE}}(x_i,y_i)$, $\forall (x_i,y_i) \in \mc{D}$.  
        }
        \Else
        {
            Obtain per-sample loss $l_i = \mc{L}_{\text{CE}}(f_p(x_i),y_i)$, $\forall (x_i,y_i) \in \mc{D}$ \\
            Compute clean probability $g(x_i,y_i)$ based on GMM\\
            Obtain $\bar{\mc{D}} = \{(x_i,y_i) | g(x_i,y_i)>p_t\}$ \\
            Train $f_p$ using $\mc{L}_{\text{CE}}(x_i,y_i)$, $\forall (x_i,y_i) \in \bar{\mc{D}}$ \\
        }
        
    }
    Compute entropy $H_{\tau}(x_i)$, $\forall (x_i,y_i) \in \mc{D}$\\
    Obtain sampling probability $\mc{P}(x_i)$, $\forall (x_i,y_i) \in \mc{D}$  \\
    \While{$e < E$\tcp*{Train robust model $f_r$}} 
    {   
        Batch sampling with $\mc{P}(x)$  \\
        Denoising on batch $\mc{A}_{\text{den}} (B)$
    }
\end{algorithm}

We summarize our pseudo-code at Section~\ref{sec:denoising} in Algorithm~\ref{alg:deneb}.

%% file: appendix/exp_setting.tex
\newpage
\section{Experimental Settings}
\label{app:exp}

\subsection{Implementation Details}
\myparagraph{Baseline Implementation.}
We reproduce all the experimental results that refer to other official repositories of denoising algorithms
\footnote{https://github.com/asappresearch/aum}\footnote{https://github.com/bhanML/Co-teaching}\footnote{https://github.com/xingruiyu/coteaching\_plus}\footnote{https://github.com/LiJunnan1992/DivideMix}\footnote{https://github.com/Kthyeon/FINE\_official}
and debiasing algorithms
\footnote{https://github.com/alinlab/LfF}\footnote{https://github.com/anniesch/jtt}\footnote{https://github.com/kakaoenterprise/Learning-Debiased-Disentangled}.
Especially, we directly inherit their loss or training method without any modification.

\subsection{Datasets}
We summarize bias benchmarks and explain how to inject noisy labels. We follow the colors in the main manuscript: \textcolor{blue}{(aligned, clean)}, \textcolor{green}{(aligned, noisy)}, \textcolor{orange}{(conflicting, clean)}, and \textcolor{red}{(conflicting, noisy)}. Because the number of bias-conflicting samples are small, we plot some of samples in the following examples.

\subsubsection{Dataset bias}
Four types of benchmarks are presented: Colored MNIST, Corrupted CIFAR, BFFHQ, and BAR. Each benchmark is derived from earlier publications~\cite{nam2020learning, kim2021biaswap, lee2021learning}. The recipe for pruducing each benchmark is detailed in below.

\begin{itemize}
\begin{figure}[h!]
    \centering
        \includegraphics[width=0.79\columnwidth]{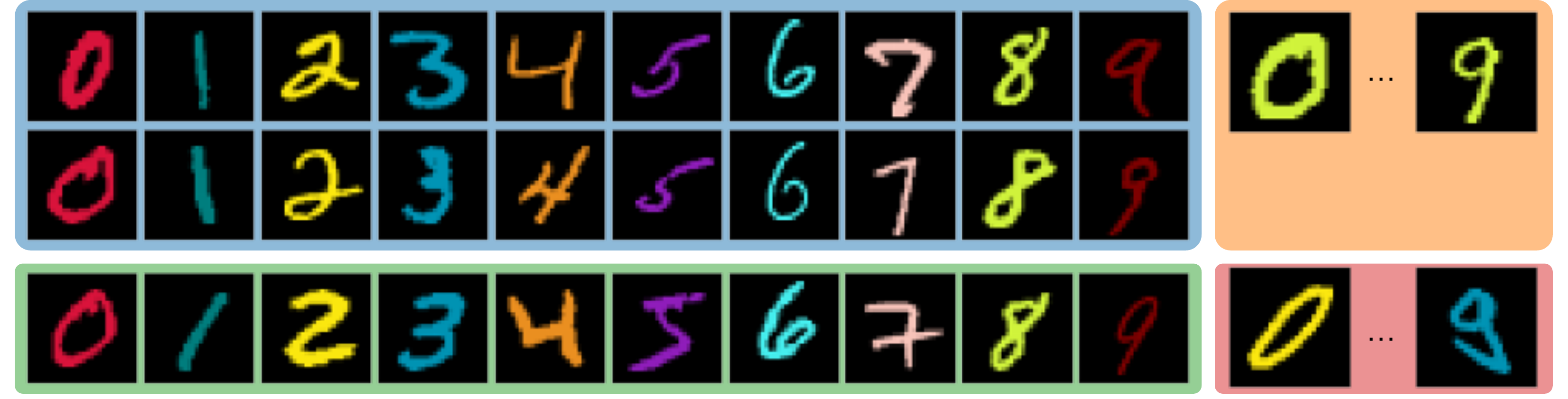}
        \caption{Colored MNIST}
        \label{appfig:cmnist}
\end{figure}

\begin{table}[!h]
    \caption{Color Summary}
    \label{tab:conflicting_summary_cmnist}
    \centering
    \resizebox{0.9\linewidth}{!}{
        \begin{tabular}{ccccccccccc}
        \thickhline
        Class & 0&1&2&3&4&5&6&7&8&9 \\ \hline
        R &  0.8627451	& 0	&  0.99215686&	0&	0.929411765	&0.568627451&	0.274509804&	0.980392157	&0.823529412&	0.501960784 \\
G &0.07843137&	0.50196078	&0.91372549	&0.58431373	&0.568627451&	0.117647059	&0.941176471&	0.77254902&	0.960784314&	0 \\
B & 0.23529412	&0.50196078	&0.0627451&	0.71372549&	0.129411765&	0.737254902	&0.941176471&	0.733333333&	0.235294118	&0 \\ 
        \thickhline
        \end{tabular}
    }
\end{table}

\item \textbf{Colored MNIST~\cite{nam2020learning, kim2021biaswap, lee2021learning}} 
Colored MNIST is a variant of the well-known handwritten digit dataset. This dataset is generated by injecting colors as follows: we pick ten different colors, $c_0,...,c_9$, and color $1-\alpha$ samples of each class with $c_i$ colors where $i=y$, and the rest are colored $c_j$ where uniformly chosen $j \neq y$. The main objective of this task is to classify the digit shape while colors are highly correlated. For injecting variance of colors, we add variance from $c_i + v \sim \mc{N}(0,0.01^2)$. Examples are described in Figure~\ref{appfig:cmnist} and the picked colors are in Table~\ref{tab:conflicting_summary_cmnist}

\begin{figure}[h!]
    \centering
        \includegraphics[width=0.79\columnwidth]{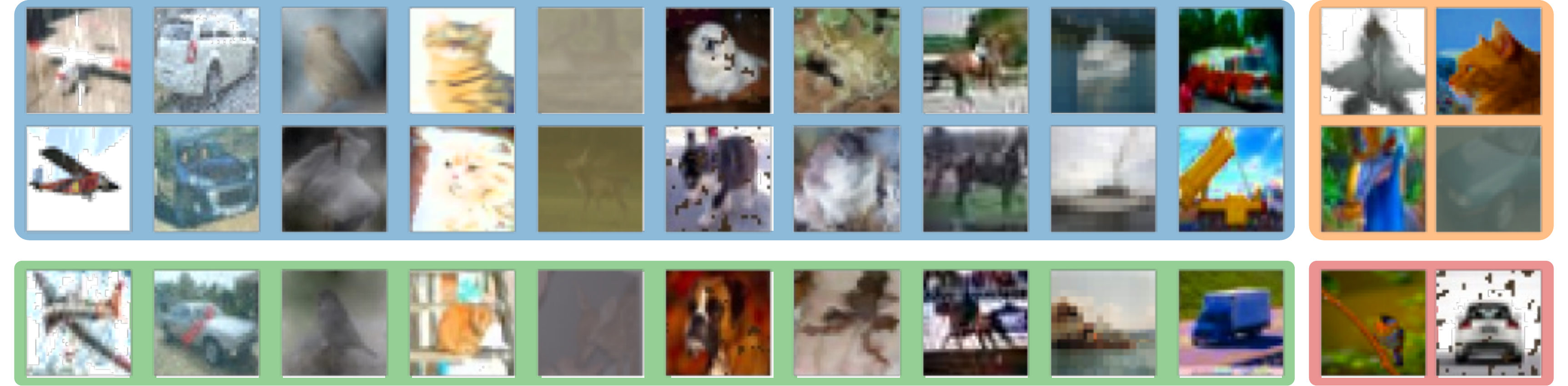}
        \caption{Corrupted CIFAR}
        \label{appfig:ccifar}
\end{figure}

\begin{table}[!h]
    \caption{Color Summary}
    \label{tab:conflicting_summary_ccifar}
    \centering
    \resizebox{0.9\linewidth}{!}{
        \begin{tabular}{ccccccccccc}
        \thickhline
        Class & 0&1&2&3&4&5&6&7&8&9 \\ \hline
        Corruption & Snow& Frost& Fog& Brightness& Contrast& Spatter& Elastic& JPEG& Pixelate& Saturate \\ 
        \thickhline
        \end{tabular}
    }
\end{table}

\item \textbf{Corrupted CIFAR~\cite{nam2020learning, lee2021learning}} This dataset comes from the conventional CIFAR dataset. In this dataset, we focus on two attributes: object and corruption. Similar to the Colored MNIST, we select corruption types, such as $c_0$=blur,..., $c_9$=snow. $1-\alpha\%$ samples of each class are corrupted with $c_i$ where $i = y$. The rest samples are blurred uniformly random samples $c_j$ where $j \neq y$. Examples are in Figure~\ref{appfig:ccifar} and the utilized corruptions are in Table~\ref{tab:conflicting_summary_ccifar}.
\begin{figure}[h!]
    \centering
        \includegraphics[width=0.79\columnwidth]{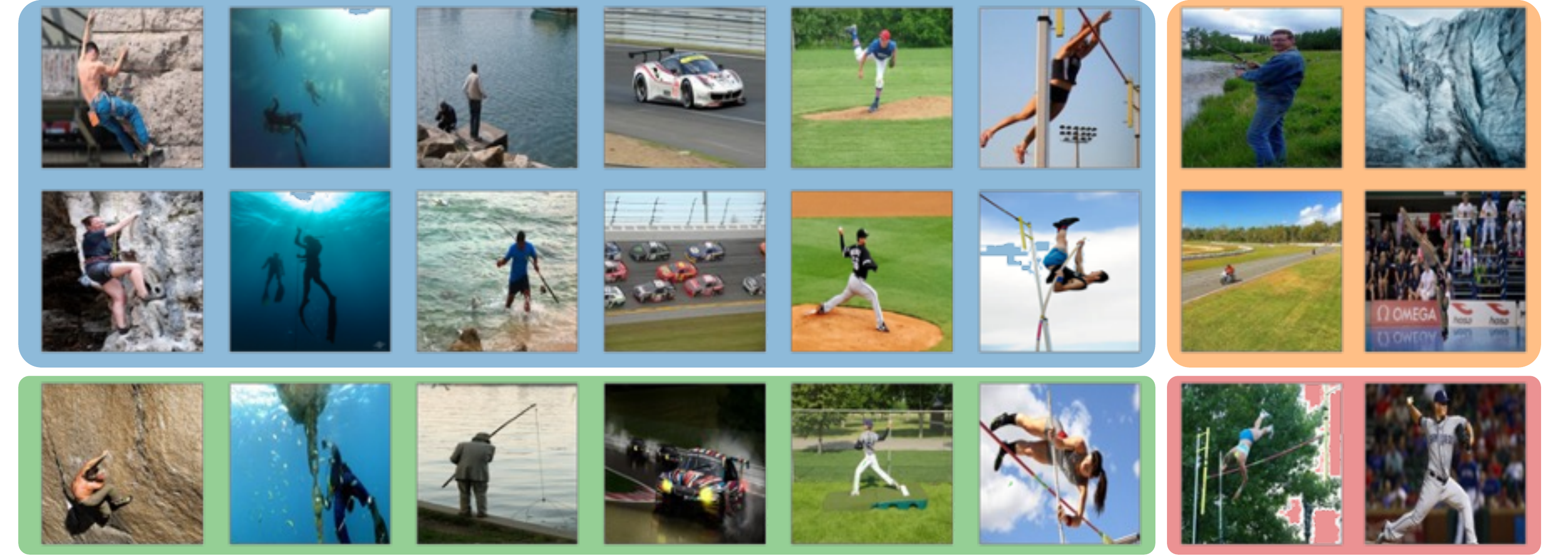}
        \caption{Biased Action Recognition}
        \label{appfig:bar}
\end{figure}
\item \textbf{BAR~\cite{nam2020learning}}\footnote{https://github.com/alinlab/LfF} BAR dataset has totally six action categoris: \eg racing, vaulting, climbing. Different with Colored MNIST and Corrupted CIFAR datasets, it does not contain manually injected bias attributes, \ie it utilizes bias raw images. This dataset has two attributes: action and background. Each class has its own biased background. For example, almost all climbing action class images are taken against cliff, while only a few samples are taken against ice-cliff.
\begin{figure}[h!]
    \centering
        \includegraphics[width=0.79\columnwidth]{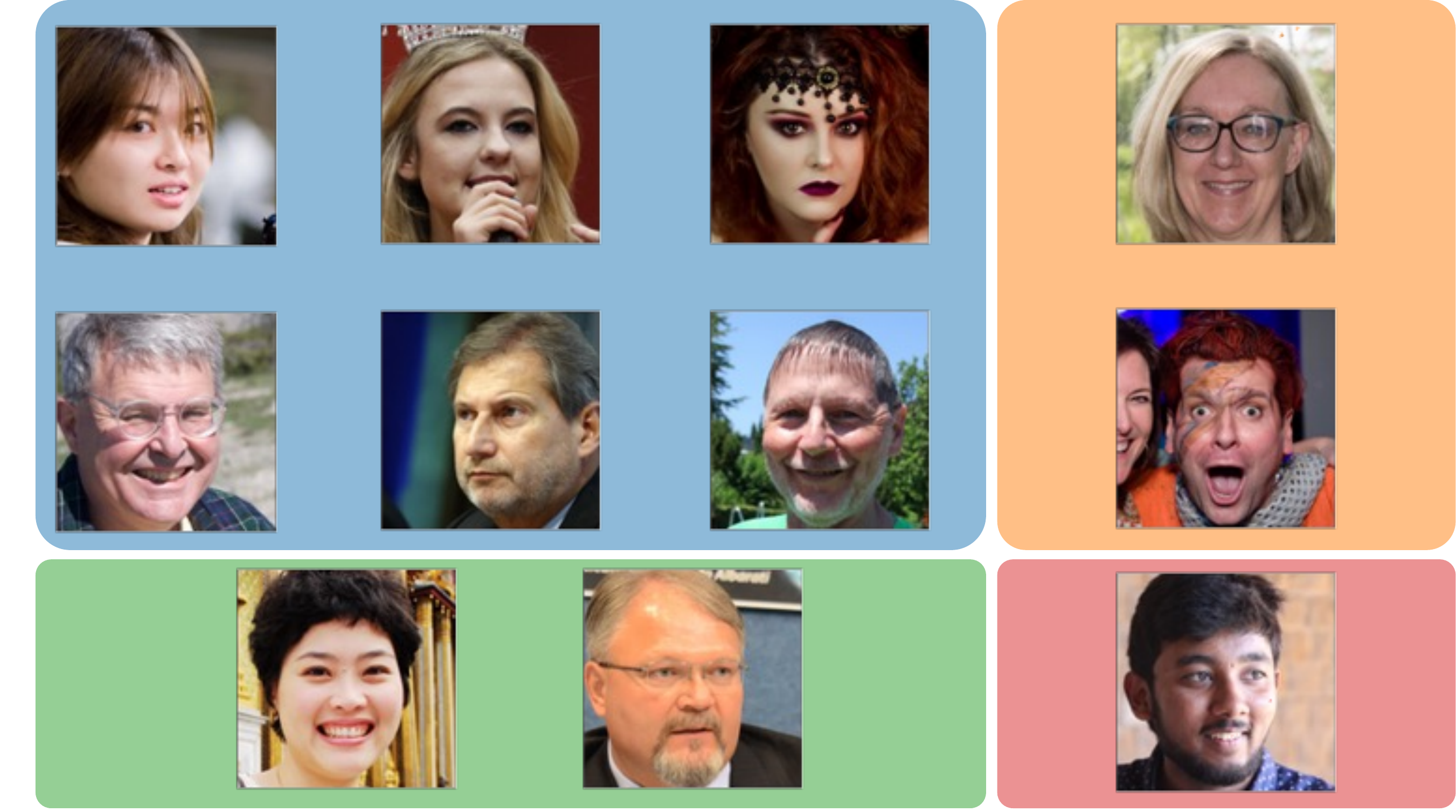}
        \caption{Biased FFHQ}
        \label{appfig:bffhq}
\end{figure}
\item \textbf{BFFHQ~\cite{kim2021biaswap, lee2021learning}}\footnote{https://github.com/kakaoenterprise/Learning-Debiased-Disentangled} BFFHQ is manually subsampled dataset from FFHQ facial images. We focus on two attributes of this dataset: gender and age. Similar with BAR dataset, we do not inject manual bias attribute. The target attribute is gender when bias the bias attribute is age. The number of classes of this benchmark is binary; \ie Male and Female.  Almost male samples are old (age ranging from 40 to 59), while female samples are young (age ranging from 10 to 29). Bias-conflicting samples are opposite; (male young) and (female old).
\end{itemize}

\subsubsection{Label flipping}
We generate noisy labels by flipping the labels symmetrically, as mentioned in the main paper. Label flipping rule is as follows: (1) Select $\eta\%$ samples (2) Select new label $y$ uniformly $\{0,...,C\}$. Note that, new label $y$ can contains ground truth label. 

\subsection{Backbone network}
\myparagraph{Simple-Conv-Net.}
We report Simple convolutional networks for Colored MNIST by \code{PyTorch} code as follows:

\noindent

\small{\code{(conv1): Conv2d(3, 8, kernel\_size=(4, 4), stride=(1, 1))\\
(bn1): BatchNorm2d(8, eps=1e-05, momentum=0.1, affine=True, track\_running\_stats=True)\\
(relu1): ReLU()\\
(dropout1): Dropout(p=0.5, inplace=False)\\
(avgpool1): AvgPool2d(kernel\_size=2, stride=2, padding=0)\\
(conv2): Conv2d(8, 32, kernel\_size=(4, 4), stride=(1, 1))\\
(bn2): BatchNorm2d(32, eps=1e-05, momentum=0.1, affine=True, track\_running\_stats=True)\\
(relu2): ReLU()\\
(dropout2): Dropout(p=0.5, inplace=False)\\
(avgpool2): AvgPool2d(kernel\_size=2, stride=2, padding=0)\\
(conv3): Conv2d(32, 64, kernel\_size=(4, 4), stride=(1, 1))\\
(relu3): ReLU()\\
(bn3): BatchNorm2d(64, eps=1e-05, momentum=0.1, affine=True, track\_running\_stats=True)\\
(dropout3): Dropout(p=0.5, inplace=False)\\
(avgpool3): AdaptiveAvgPool2d(output\_size=(1, 1))\\
(fc): Linear(in\_features=64, out\_features=\textbf{\$num\_class}, bias=True)
}}

\myparagraph{ResNet18 pre-trained on ImageNet.} Except for Colored MNIST benchmark, we use pretrained ResNet18. This network is known that pre-trained using ImageNet. We obtain the parameters from torchvision library.

\newpage

\input{table/hyperparam}

\input{table/hyper_ours}

\subsection{Hyperparameter tuning}
Under our experiment, we searched best model under following search space in Table~\ref{tab:baseline-specific}. Common hyperparameters, such as Learning rate, Batch size, LR decay rate, are shared for all algorithms. Since all algorithms have different hyper-parameters, we tune with $30$ trials for fair comparison. Hyperparameter tuning was conducted based on \code{ASHAScheduler}, efficient hyperparameter tuning module supported by official \code{Ray}~\cite{liaw2018tune}. We select the best validation accuracy model among all candidates.

\myparagraph{Implementation details for~\alg}
We summarize the hyperparameters for training~\alg~for each benchmark. Detail parameters are in Table~\ref{tab:our_hyper}. These parameters are also obtained by following the hyper-parameter tuning rule.

%% file: table/hyperparam.tex
\begin{table}[t!]
    \caption{Hyperparameter search spaces}
    \label{tab:baseline-specific}
    \centering
    \resizebox{0.45\columnwidth}{!}{
        \begin{tabular}{ccc}
        \thickhline
        Algorithm & Name & Candidate \\ \hline
        \multirow{6}{*}{Common} & LR                    & $\{0.1, 0.01, 0.001, 0.0001\}$         \\
                                & Batch size            & $\{32, 64, 128, 256\}$  \\ 
                                & LR Decay              & $\{1, 0.1 ,0.01 ,0.001, 0.0001\}$        \\ 
                                & Lr decay step         & $\{20, 30, 40\}$        \\ 
                                & Momentum              & $\{0.1, 0.3, 0.5, 0.7, 0.9\}$          \\ 
                                & weight decay          & $\{0.0, 0.1, 0.01, 0.001 ,0.0001\}$         \\ \hline
        \multirow{2}{*}{\code{LfF}} & $q$          & $\{0.1, 0.5, 0.7, 0.9\}$         \\
                                    & $\alpha$          & $\{0.01, 0.1, 0.5, 0.7, 0.9, 0.99\}$          \\ \hline
        \multirow{2}{*}{\code{JTT}} & bias epochs          & $\{1, 10, 20, 50, 100\}$          \\
                                    & $\lambda_{up}$          & $\{1., 10., 20., 50., 100.\}$          \\ \hline
        \multirow{5}{*}{\code{Disen}} & $\alpha$          & $\{0.01, 0.1, 0.5, 0.7, 0.9, 0.99\}$          \\
                                    & $q$          & $\{0.01, 0.1, 0.5, 0.7, 0.9, 0.99\}$          \\ 
                                    & warmup          & $\{0, 10, 20\}$          \\ 
                                    & $\lambda_{dis}$          & $\{1, 10, 30\}$          \\ 
                                    & $\lambda_{swap}$          & $\{1, 10, 30\}$          \\ \hline
        \multirow{2}{*}{\code{SCE}} & $\alpha$          & $\{0.01, 0.1, 0.5, 0.7, 0.9, 0.99\}$          \\ 
                                    & $\beta$          & $\{1.0, 0.5, 0.7\}$          \\ \hline
        \multirow{1}{*}{\code{GCE}} & $q$          & $\{0.1, 0.3, 0.5, 0.7, 0.9\}$          \\ \hline
        \multirow{7}{*}{\code{ELR+}} & $\lambda$          & $\{1, 3, 5, 7\}$          \\ 
                                    & $\beta$          & $\{0.7, 0.9\}$          \\ 
                                    & warmup          & $\{0, 10, 30\}$          \\ 
                                    & ema $\alpha$          & $\{0.9, 0.99, 0.997\}$          \\ 
                                    & ema step         & $\{30000, 40000, 50000\}$          \\ 
                                    & mixup $\alpha$          & $\{0.5, 1.0, 2.0, 5.0\}$          \\ 
                                    & coef step          & $\{0, 100, 1000\}$          \\  \hline
        \multirow{1}{*}{\code{AUM}} & Percentile          & $\{1, 10, 30, 50, 99\}$          \\  \hline
        \multirow{2}{*}{\code{Co-teaching}} & num gradual          & $\{0, 2, 5, 10\}$          \\ 
                                    & forget rate          & $\{0.1, 0.3, 0.5, 0.9\}$          \\  \hline
                                    
        \multirow{5}{*}{\code{DivideMix}} & $\alpha$          & $\{0.5, 1.0, 2.0, 5.0\}$          \\ 
                                    & $\lambda_u$          & $\{0, 0.5, 1, 15, 25\}$          \\ 
                                    & $p_{threshold}$          & $\{0.1, 0.3, 0.5, 0.7\}$          \\ 
                                    & warmup          & $\{0, 5, 10\}$          \\ 
                                    & $T$          & $\{0.1, 0.5\}$          \\  \hline
        \multirow{5}{*}{\code{f-DivideMix}} & $\alpha$          & $\{0.5, 1.0, 2.0, 5.0\}$          \\ 
                                    & $\lambda_u$          & $\{0, 0.5, 1, 15, 25\}$          \\ 
                                    & $p_{threshold}$          & $\{0.1, 0.3, 0.5, 0.7\}$          \\ 
                                    & warmup          & $\{0, 5, 10\}$          \\ 
                                    & $T$          & $\{0.1, 0.5\}$          \\  \hline
        \multirow{4}{*}{\code{DENEB}} & $p_{threshold}$          & $\{0.1, 0.3, 0.5, 0.7, 0.9\}$          \\ 
                                    & warmup          & $\{0, 5, 10\}$          \\ 
                                    & $\tau$          & $\{0.1, 0.5, 0.9, 1.0, 2.0, 5.0, 10.0\}$          \\   \hline
        \thickhline
    \end{tabular}}
\end{table}

%% file: table/hyper_ours.tex
\begin{table}[t!]
    \caption{Hyperparameters for~\alg}
    \label{tab:our_hyper}
    \centering
    \resizebox{0.8\columnwidth}{!}{
        \begin{tabular}{ccccccc}
        \thickhline
        Benchmark & Type & $p_{threshold}$ & $\tau$ & $e_w$ & $q$ & Augmentation \\ \hline
        CMNIST & $(\alpha = 1\%, \eta = 10\%)$ & 0.1 & 1.0 & 5 & 0.5 &  N/A       \\  
        CMNIST & $(\alpha = 5\%, \eta = 50\%)$ & 0.9 & 2.0 & 5 & 0.7 &  N/A \\ 
        CCIFAR & $(\alpha = 1\%, \eta = 10\%)$ & 0.9 & 10.0 & 5 & 0.3 & RandomCrop(32,padding=4),RandomHorizontalFlip, Normalize((0.4914, 0.4822, 0.4465), (0.2023, 0.1994, 0.2010))             \\  
        CCIFAR & $(\alpha = 5\%, \eta = 50\%)$ & 0.7 & 10.0 & 10 & 0.3 & RandomCrop(32,padding=4),RandomHorizontalFlip, Normalize((0.4914, 0.4822, 0.4465), (0.2023, 0.1994, 0.2010))             \\ 
        BAR & $(\eta = 10\%)$ & 0.5 & 1.0 & 10 & 0.7 & Normalize((0.4914, 0.4822, 0.4465), (0.2023, 0.1994, 0.2010))         \\  
        BFFHQ & $(\eta = 10\%)$ & 0.5 & 1.0 & 10 & 0.7 &RandomCrop(128,padding=4),Normalize((0.4914, 0.4822, 0.4465), (0.2023, 0.1994, 0.2010))         \\  
        \thickhline
    \end{tabular}}
\end{table}

%% file: appendix/obs_setting.tex
\newpage
\section{Experimental Settings for Section~\ref{sec:debiasing} and~\ref{sec:denoising}}
\label{app:obs_setting}

\subsection{Implementation for Section~\ref{sec:debiasing}}
To see the impact of the noisy labels, we observe how peformances are degraded when the labels are corrupted, as in Section~\ref{sec:debiasing}. 

\myparagraph{Label-based implementation.}
To check how the label-based approaches performs when noisy labels are injected, we test three baselines~\cite{nam2020learning, lee2021learning, liu2021just}.
For all baselines, we clone the official codes from their \code{github} site. Also, we download datasets as instructed in the site. After then, we only modify the dataloader part of the code for injecting noisy labels. Hyperparameters and running commands are exactly followed as in the instructions of all algorithms. 

\myparagraph{Label-free implementation.}
To see the impact of label-free method, we utilize entropy score. Here, we summarize our entropy based approach which is composed of three steps. First, we train the biased model based on GCE with parameter $q = 0.7$. This is similar with \code{LfF} and \code{Disen}. After fully train the biased model, we extract per-sample entropy. Based on the computed entropy, we obtain sampling probability of each sample, and run the stage 3 of~\alg.

\subsection{Implenetation for Section~\ref{sec:denoising}}
To see the impact of denoising on DBwNL dataset, we run five denoising algorithms. Except for \code{AUM}, all algorithms are online-manner. Therefore, we report the cleansing result at the end of training phase. For example, when we run the \code{DivideMix} with $100$ epochs, we extract division results after training $100$ epoch.  Furthermore, to see the case when we can emphasize bias-conflicting samples, we multiply $\times 50$ for the bias-conflicting samples, regardless of it is noisy labels or not.

\subsection{Other case study in Section~\ref{sec:debiasing} and~\ref{sec:denoising}}
\begin{figure}[h!]
    \centering
    \includegraphics[width=0.3\linewidth]{./fig/histogram_legend_two.pdf}
    \centering
    \subfloat[\code{LfF}~\cite{nam2020learning}]{\label{fig:lff_h}\includegraphics[width=0.24\linewidth]{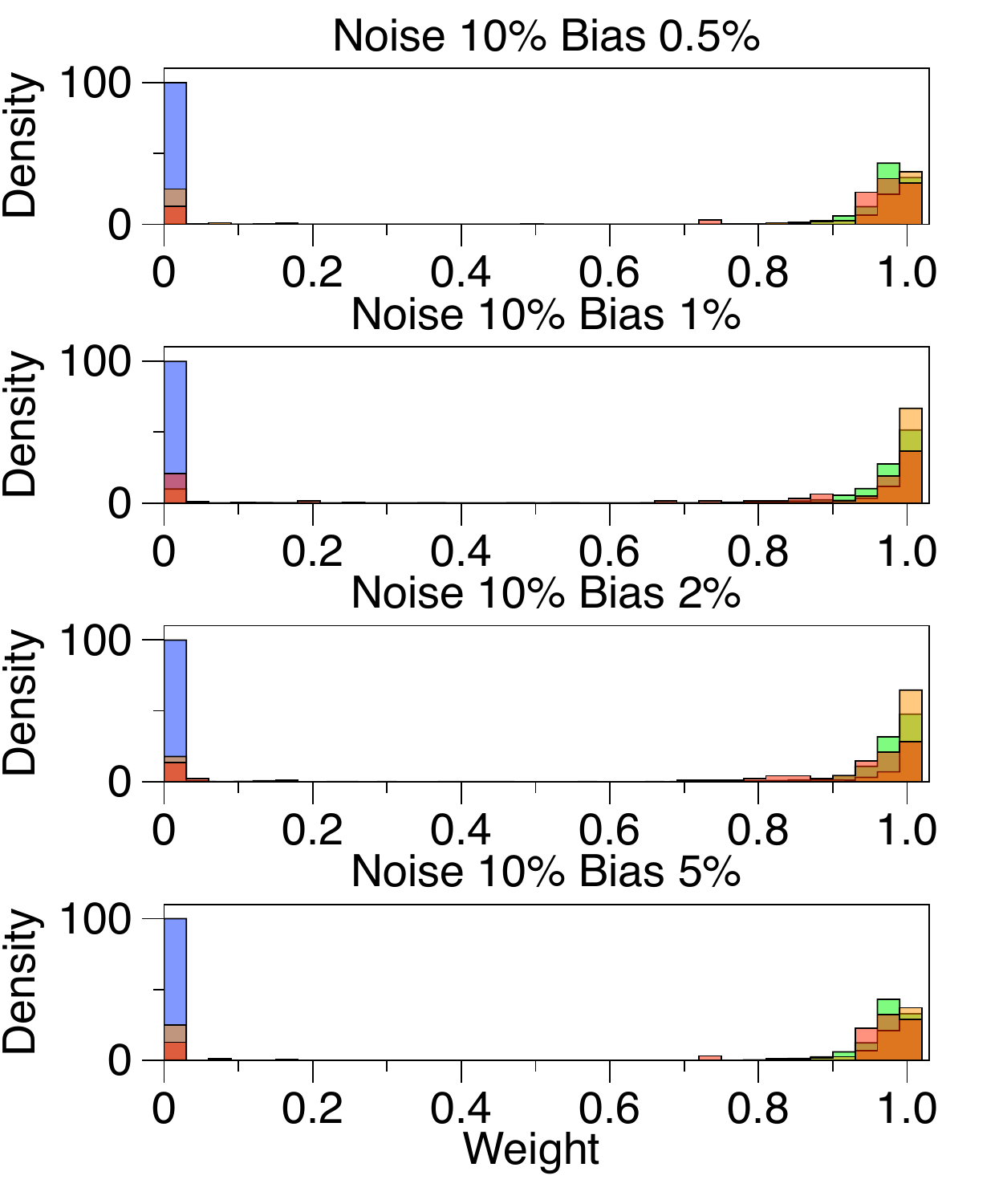}}
    \subfloat[\code{JTT}~\cite{liu2021just}]{\label{fig:jtt_h}\includegraphics[width=0.24\linewidth]{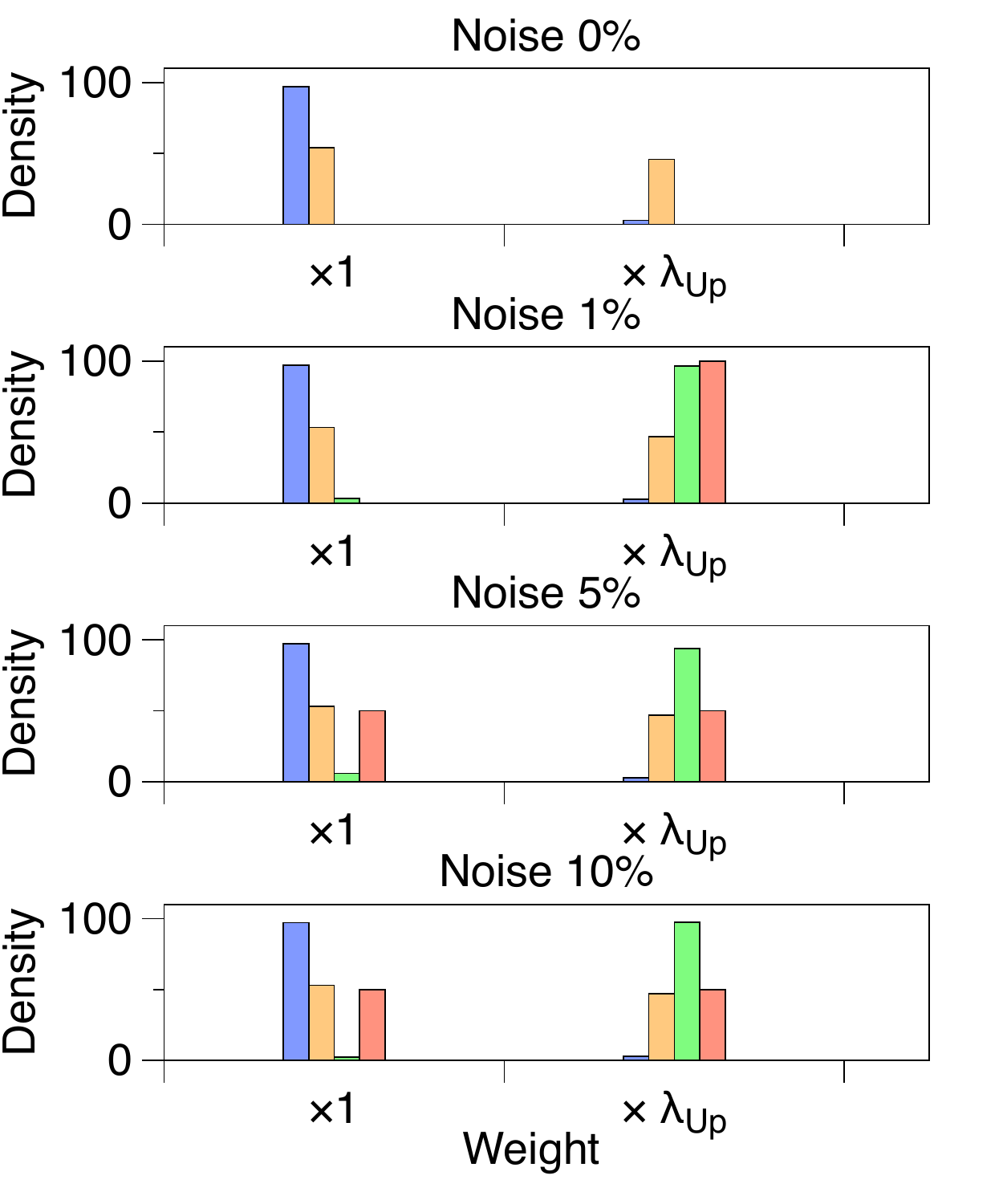}}
    \subfloat[\code{Disen}~\cite{lee2021learning}]{\label{fig:disen_h}\includegraphics[width=0.24\linewidth]{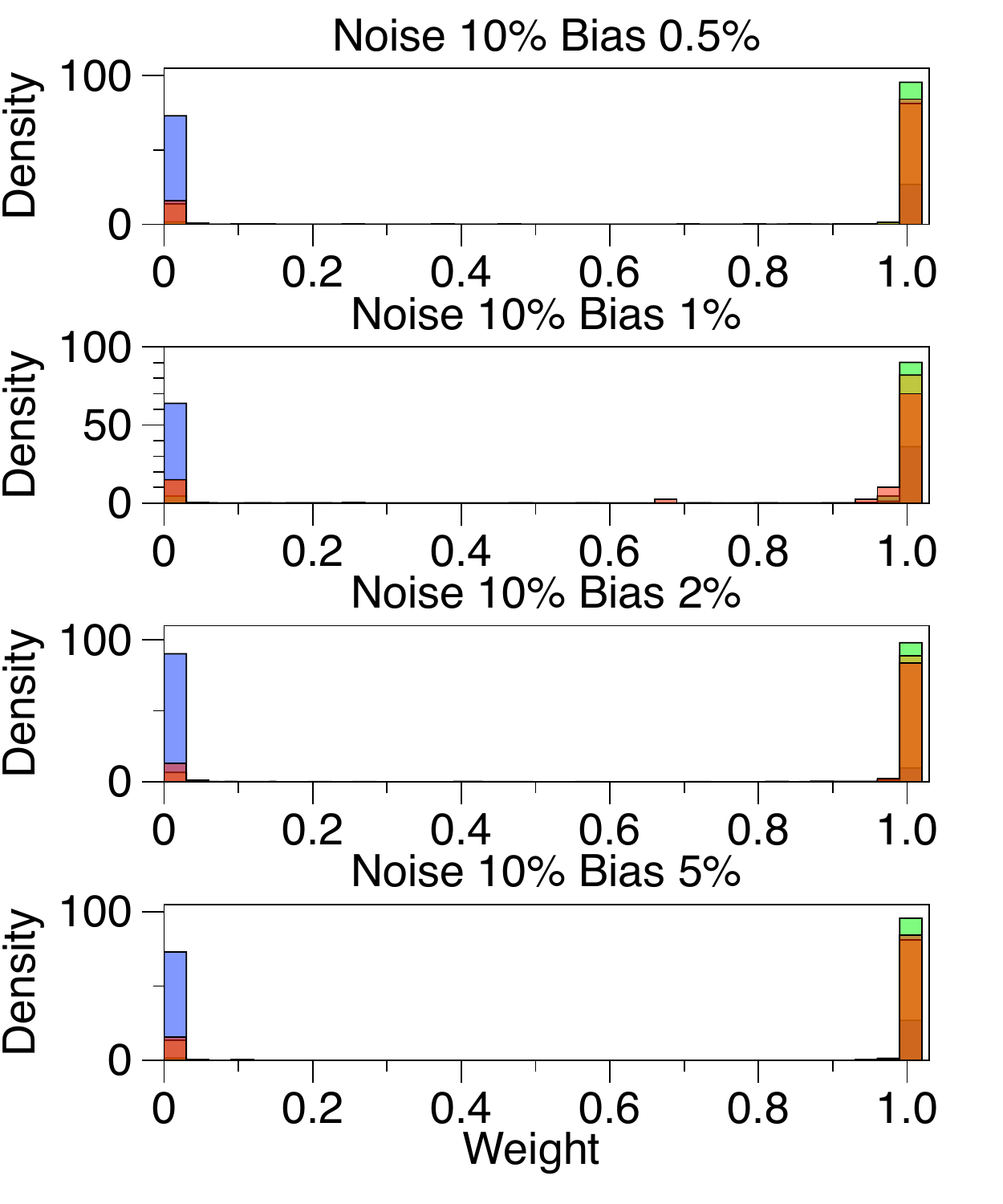}}
    \subfloat[\code{Entropy}]{\label{fig:entropy_h}\includegraphics[width=0.24\linewidth]{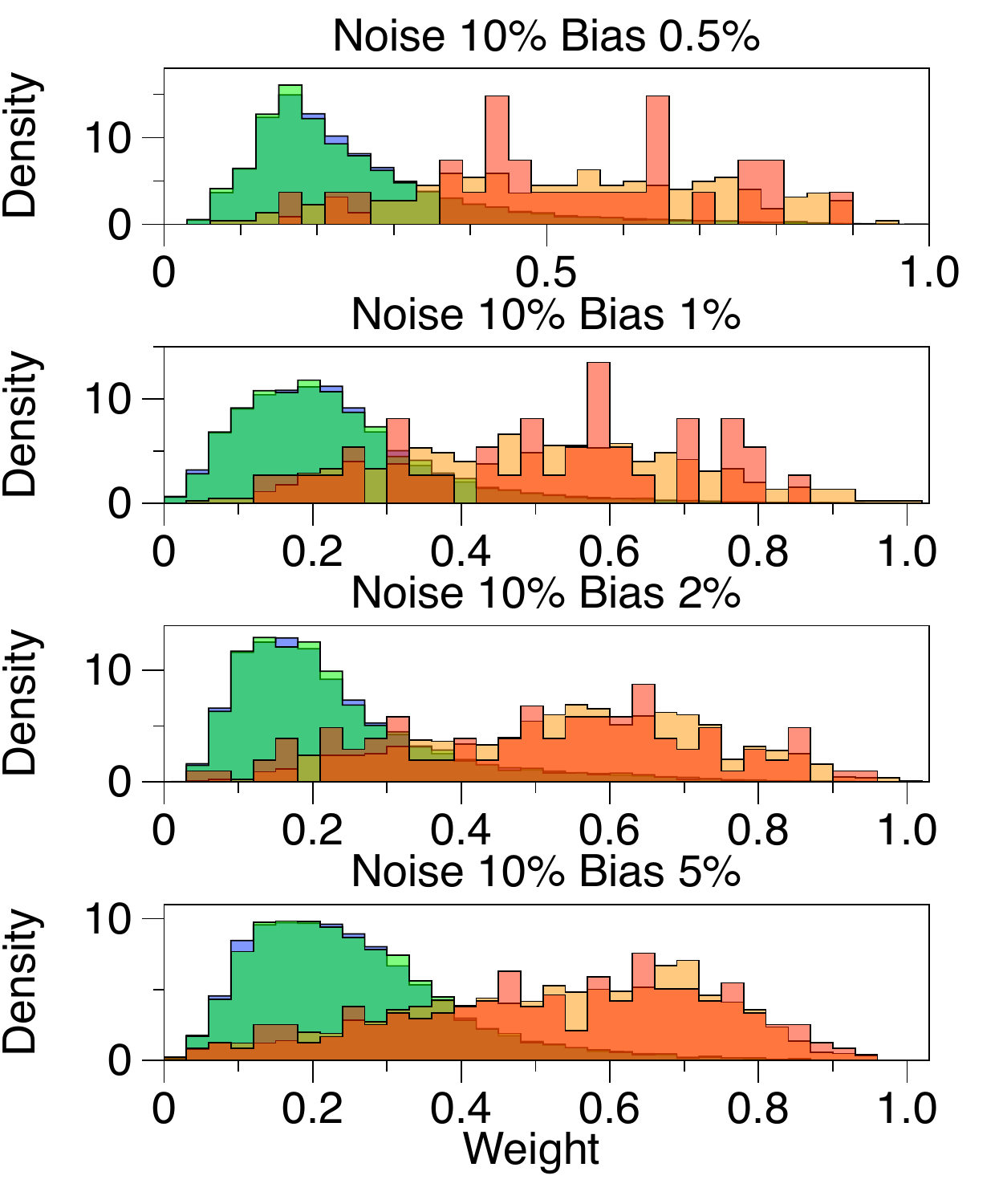}}
    \caption{Score histogram of each methodology. As \code{LfF} and \code{Disen} operate online, we report the weight histogram right after the last epoch of training. Except for the Entropy case, \code{LfF}, \code{JTT}, and \code{Disen} shows entangled histograms between \textcolor{green}{(noisy, aligned)}, \textcolor{orange}{(clean, conflicting)}, and  \textcolor{red}{(noisy, conflicting)}. By contrast, the histogram of entropy case indicates that it is clustered not according to label corruption but bias.}
    \label{fig:histogram_appendix}
\end{figure}

To verify our findings, we additionally offer the other cases. We change $\alpha$ and $\eta$. As in Figure~\ref{fig:histogram_appendix}, previous label-based approaches have entangled

\begin{figure}[h!]
    \centering
    \includegraphics[width=0.3\textwidth]{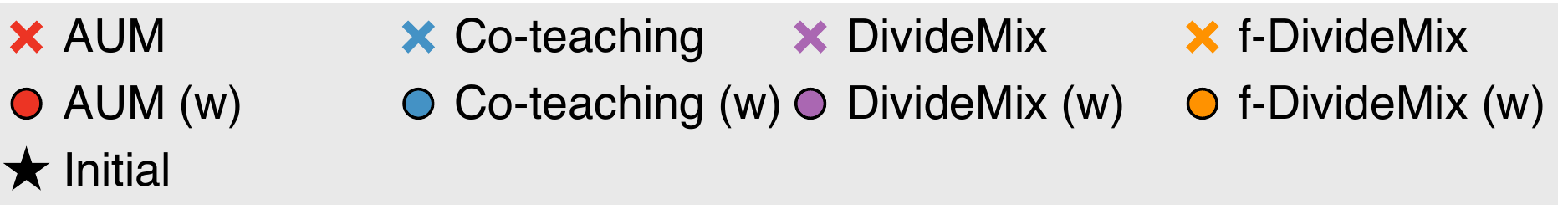}\\
    \centering
    \subfloat[$\alpha = 1\%$, $\eta = 10\%$]
    {\label{fig:denoise_1_10}\includegraphics[width=0.24\linewidth]{./fig/Bias_1_noise_10.pdf}}
    \subfloat[$\alpha = 1\%$, $\eta = 50\%$]
    {\label{fig:denoise_1_50}\includegraphics[width=0.24\linewidth]{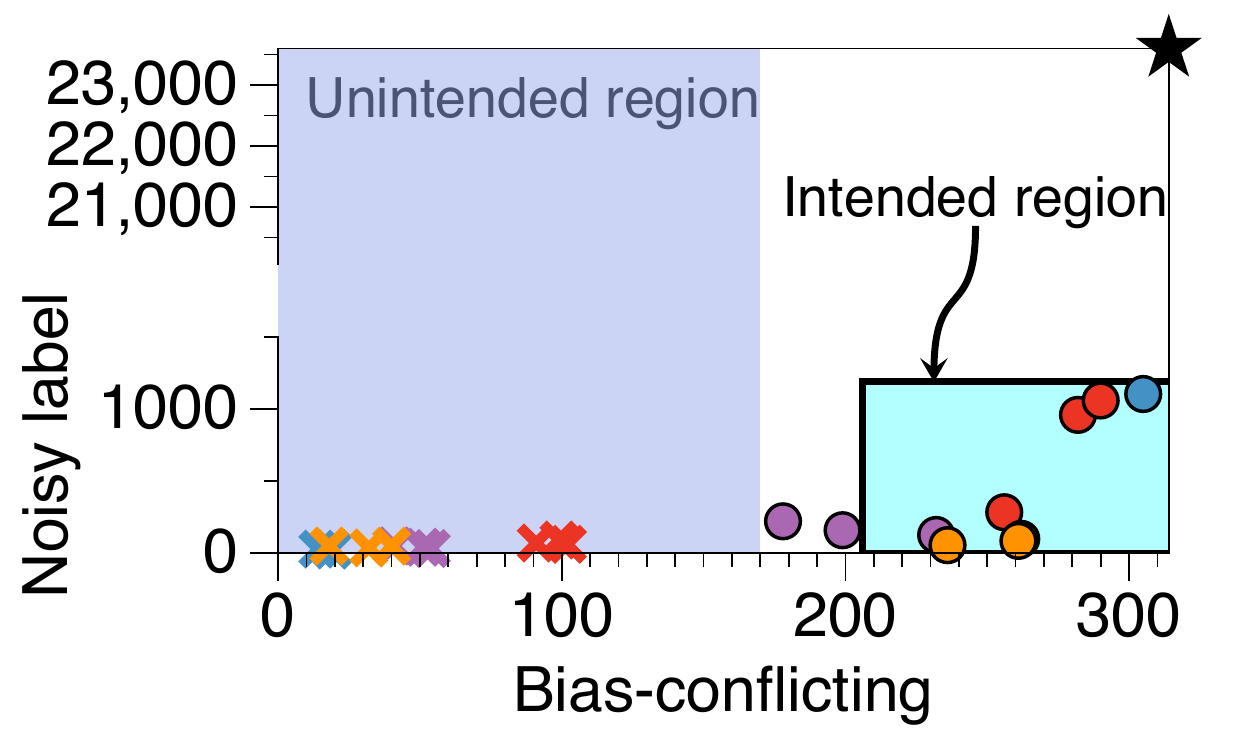}}
    \subfloat[$\alpha = 5\%$, $\eta = 10\%$]
    {\label{fig:denoise_5_10}\includegraphics[width=0.24\linewidth]{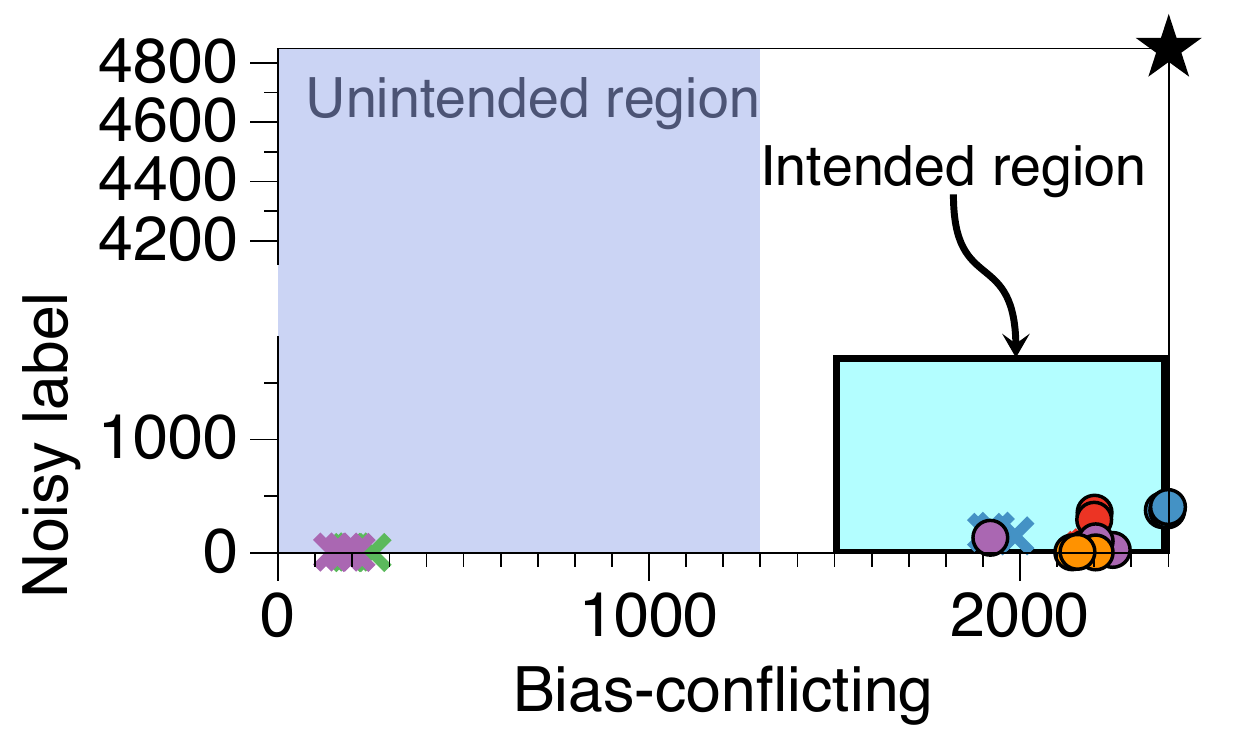}}
    \subfloat[$\alpha = 5\%$, $\eta = 50\%$]
    {\label{fig:denoise_5_50}\includegraphics[width=0.24\linewidth]{./fig/Bias_5_noise_50.pdf}}
    \caption{Number of remaining noisy labels and bias-conflicting samples after denoising is conducted. Star $\star$ mark represents the number of samples before cleansing, and \xmark~ and $\CIRCLE$ marks indicate with or without weighted training results. Since bias-conflicting samples is precious for debiasing, bias-conflicting samples have to be protected. Therefore, the region loses bias-conflicting samples (left, blue) is the unintended region. On the other hand, the region ignores noisy labels without losing the bias-conflicting samples (right, cyan) is the intended behavior.}
    \label{fig:denoise_appendix}
\end{figure}

%% file: appendix/computing.tex
\newpage
\section{Computation Resources}
\label{app:compute_resource}
For all experiments, we utilized \code{NVIDIA TITAN Xp} GPUs and \code{Intel Xeon CPU E5-2630 v4} CPU cores. $252$Gb RAM memory was available but, we partially utilized it. For Colored MNIST and Corrupted CIFAR cases, we use only one GPU. On the other hand, we use 4 GPUs for BAR and BFFHQ benchmarks. We summarize training time of CMNIST for each entity in Table~\ref{tab:training_time}.

\begin{table}[!h]
    \caption{Color Summary}
    \label{tab:training_time}
    \centering
    \resizebox{0.3\linewidth}{!}{
        \begin{tabular}{cccc}
        \thickhline
        Step & Step 1& Step 2& Step3 \\ \hline
        Training time & 9m 3s & 5s & 6m 12s \\ 
        Vanilla & \multicolumn{3}{c}{5m 59s} \\
        \thickhline
        \end{tabular}
    }
\end{table}